\documentclass{article} % For LaTeX2e
\usepackage{iclr2024_conference,times}

\usepackage{amsmath,amsfonts,bm}

\def\eqref#1{equation~\ref{#1}}
\def\1{\bm{1}}

\DeclareMathAlphabet{\mathsfit}{\encodingdefault}{\sfdefault}{m}{sl}
\SetMathAlphabet{\mathsfit}{bold}{\encodingdefault}{\sfdefault}{bx}{n}

\usepackage{hyperref}
\usepackage{url}
\usepackage{epsfig}
\usepackage{graphicx}
\usepackage{amsmath}
\usepackage{amssymb}
\usepackage{subcaption}

\usepackage{subfloat}
\usepackage{appendix}
\usepackage{multirow}
\usepackage{float}
\usepackage{booktabs}
\usepackage{color}
\usepackage{tikz}
\usepackage{graphicx}

\usepackage{tcolorbox}
\tcbuselibrary{most}
\title{Benchmarking Sequential Visual Input Reasoning and Prediction in Multimodal Large Language Models}

\author{
Mingwei Zhu\textsuperscript{1}, Leigang Sha\textsuperscript{1}, Yu Shu\textsuperscript{1}, 
Kangjia Zhao\textsuperscript{1}, Tiancheng Zhao\textsuperscript{2,*}, Jianwei Yin\textsuperscript{1} \\
\textsuperscript{1}Zhejiang University \quad 
\textsuperscript{2}Binjiang Institute of Zhejiang University \\
\texttt{\{zhumw, shaleigang, shu\_yu, konkaz, zjuyjw\}@zju.edu.cn} \\
\texttt{tianchez@zju-bj.com}
}

\iclrfinalcopy % Uncomment for camera-ready version, but NOT for submission.

\begin{document}
\maketitle
\begin{abstract}
Multimodal large language models (MLLMs) have shown great potential in perception and interpretation tasks, but their capabilities in predictive reasoning remain under-explored. To address this gap, we introduce a novel benchmark that assesses the predictive reasoning capabilities of MLLMs across diverse scenarios. Our benchmark targets three important domains: abstract pattern reasoning, human activity prediction, and physical interaction prediction. We further develop three evaluation methods powered by large language model to robustly quantify a model's performance in predicting and reasoning the future based on multi-visual context. Empirical experiments confirm the soundness of the proposed benchmark and evaluation methods via rigorous testing and reveal pros and cons of current popular MLLMs in the task of predictive reasoning. Lastly, our proposed benchmark provides a standardized evaluation framework for MLLMs and can facilitate the development of more advanced models that can reason and predict over complex long sequence of multimodal input. Our code is available at \url{https://github.com/CoderJ-ONE/Giraffe-Bench}.
\end{abstract}

\section{Introduction}
Multimodal Large Language Models (MLLMs)~\cite{OpenAI2023GPT4TR} have become a crucial area of research due to their huge potential in many applications, such as Autonomous Driving~\cite{yurtsever2020survey}, Embodied Agents~\cite{driess2023palm,wu2023plan}, Robotics~\cite{ahn2022can,brohan2023rt}, and many others. Current MLLMs use text-to-image pairing data to convert visual inputs into visual tokens and integrate them into existing LLMs~\cite{liu2023visual}. Current evaluation also focuses on the perception and reasoning ability of MLLMs given an image, such as image captioning~\cite{vinyals2016show}, visual question answering~\cite{antol2015vqa}, and so on. However, in many applications, MLLMs are required to reason over a sequence of visual inputs, such as the last five frames of video input to decide a robot's next action~\cite{brohan2023rt}. This raises the question of whether or not current MLLMs, which are mostly trained on single image-text pairs, emerge the ability to reason over multiple image inputs and predict what's coming next given the context information.

Therefore, this work proposes a novel benchmark to quantify the predictive reasoning ability over three important scenarios: abstract pattern reasoning, human activity prediction, and physical interaction prediction. These three distinctive tasks require an MLLM to equip with both complex reasoning over multiple visual inputs and also use common sense world knowledge to generate highly probable outcomes. Another challenge addressed is how to robustly quantify MLLMs' predictive reasoning ability. We utilize text-only Large Language Model as a referee model and construct three paradigms of evaluation, i.e., tasks with single-golden answer, tasks with multiple-golden answers, and tasks with probabilistic outcomes.

The proposed benchmark and evaluators are tested with six popular MLLMs. Testing results confirm the soundness of the proposed tasks and evaluators. Experiment results also show surprisingly different outcomes compared to many other MLLM benchmarks, such as MME~\cite{fu2023mme}. The authors show that simple models such as LLaVA~\cite{liu2023visual} show the strongest generalization for predictive reasoning. They also show a huge gap between upper bound performance and current MLLMs' ability, shining lights on the development of more advanced MLLMs in the future. In summary, this paper's contributions are:

\begin{enumerate}
\item We propose three challenging tasks to test the predictive reasoning of MLLMs and develop high-quality datasets for model evaluation.
\item We propose three novel evaluation methods based on large language models to quantify the predictive reasoning capabilities of MLLMs.
\item We present experiment results that validate the proposed evaluators, producing evaluation scores that highly correlate with an MLLM's predictive reasoning abilities.
\item We provide benchmark results that reveal the pros and cons of current popular MLLMs, providing insights on various MLLM network structures and potential improvement directions.
\end{enumerate}

\section{Related work}
\label{headings}
\subsection{Multimodal Large Language Models}
Multimodal Large Language Models (MLLMs) are large language models~\cite{brown2020language} that can accept multimodal inputs, such as images. Notable work inlcudes LLaVA~\cite{liu2023visual} that combines CLIP~\cite{radford2021learning} and Vicuna~\cite{vicuna2023} with simple linear adapters and finetune on multimodal instruction tuning data. Later, unlike plain vision transformer patch features in LLava, MiniGPT-4~\cite{zhu2023minigpt} employs a more sophisticated QFormer~\cite{li2023blip} architecture to extract condensed visual token features as input to large language models. InstructBLIP~\cite{dai2023instructblip} further augment the training data with 13 diverse tasks including video data to improve the instruction following ability of MLLMs. mPLUG-Owl~\cite{ye2023mplug} corrects the alignment errors and enhance its multi-turn dialogue capabilities. Conversely, Otter~\cite{li2023otter} uses cross-attention architecture~\cite{awadalla2023openflamingo} for more fine-grained vision-language fusion. Lastly, Lynx~\cite{zeng2023matters}, capable of inferring on video-level sequences, adopts prefix-tuning \cite{li2021prefix} for instruction tuning \cite{peng2023instruction}, presenting a more efficient alternative to cross-attention formats.

% Amid the rapid evolution of decoder-only causal language models, the focal point in multimodal research is shifting towards enhancing the mapping between visual data and the latent space of language decoders. From the open-source LLaMA \cite{touvron2023llama} to the refined Vicuna, language models have exhibited outstanding performance in essential traditional NLP tasks such as machine translation, syntactic analysis, and other foundational areas.
%Different from the previous models, mPLUG-Owl~\cite{ye2023mplug} effectively corrects alignment errors triggered by the freezing of the visual encoder during the fine-tuning phase, enhancing its multi-turn dialogue capabilities. Conversely, Otter \cite{li2023otter}, employing the unique Open Flamingo \cite{awadalla2023openflamingo} architecture, introduces a unique visual parser. It seamlessly integrates cross-attention layers within the language model employing an in-context learning \cite{dong2022survey} training strategy, further refining the model by learning from contextual video sequences.

%Lastly, Lynx \cite{zeng2023matters}, capable of inferring on video-level sequences, adopts prefix-tuning \cite{li2021prefix} for instruction tuning \cite{peng2023instruction}, presenting a more efficient alternative to cross-attention formats. It pioneers in introducing an open-ended visual question-answering test set, transcending multiple-choice probability questions, and marking a significant advancement in the continuous enhancement of multimodal research.

\subsection{Future Prediction based on Video Input}
Video prediction is a significant area of research in computer vision. One example is the Robotic Pushing dataset~\cite{finn2016unsupervised}, which contains 59,000 robot interactions focused on pushing actions. This dataset enables precise video prediction by using the robot's impending actions as a conditioning factor. Another example~\cite{liang2019peeking} studies human activity prediction based on surveillance camera videos. CLEVRER dataset~\cite{yi2020clevrer}, which uses a 3D engine to simulate physical object motion and predicts collision effects based on trajectory paths. SUTD-TrafficQA~\cite{xu2021sutd} dataset explores real-world scenarios from a first-person perspective, assessing the possibility of car accidents within the driver's field of vision. In the realm of autonomous driving, the nuScenes dataset~\cite{caesar2020nuscenes} leverages urban traffic data to evaluate trajectory predictability between pedestrians and vehicles in busy city traffic environments.

\subsubsection{Multimodal Large Language Model Evaluation}
Despite of the impressive multimodal understanding abilities of MLLMs, they also face the difficulties of evaluation due to their generative nature. VL-Checklist~\cite{zhao2022vl} proposes to evaluate multimodal foundation models fine-grained performance to recognize the existence of visual element given complex text prompts. Mimic-it~\cite{li2023mimic} introduces a 2.8 million multimodal instruction-response dataset to bolster the zero-shot performance of multimodal models. MME~\cite{fu2023mme} establishes a comprehensive evaluation framework focusing on perception and cognition. MMBench~\cite{liu2023mmbench} enhances evaluation robustness by translating free-form answers into predefined choices. Finally, SeedBench\cite{li2023seed} specializes in generative comprehension, featuring a significantly larger human-annotated multiple-choice dataset. None of prior work has focused on studying MLLM's ability to predict and reason over sequential visual input, which is the key contribution of this paper.

\section{Proposed Benchmark and Methods}
\label{headings}
We introduce three tasks, each with its own set of multiple datasets, focusing on \textit{abstract patterns reasoning}, \textit{human activity prediction}, and \textit{physical interaction prediction}. The abstract patterns reasoning task is used to examine a model's raw reasoning ability; the human activity prediction task is used to measure a model's understanding of human intention and physical interaction prediction is used to test a model's grasp of physical laws.

% The challenges inherent in each task are diverse. Abstract patterns demand cognitive abstraction and summarization, requiring the MLLMs to introspect and synthesize complex visual patterns. Human daily activities pose challenges in subjectivity and variability, necessitating a broad understanding of social norms and daily life common sense for precise predictions. Physical interaction tasks require keen perception of motion trajectories and an understanding of physical laws, including the ability to predict missing trajectory frames from contextual cues.

% To navigate these challenges, we employ tailored evaluation metrics for each task. Our findings, fortified by ablation studies, show variable performance across these tasks, underlining the significant challenges they pose to the current capabilities of MLLMs. 
\begin{figure}
  \centering
  \includegraphics[scale=0.11]{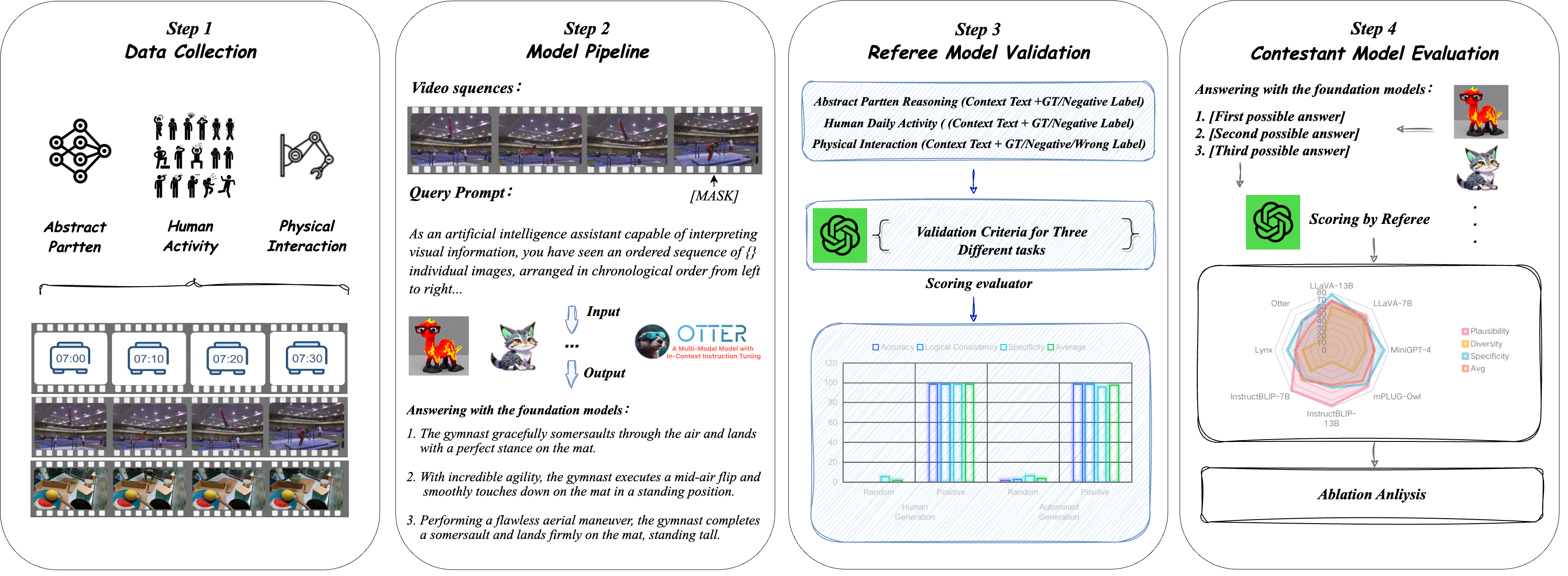}
  \caption{Overview of the proposed benchmark and method for MLLM evaluation. }
\end{figure}

\subsection{Sequential Visual Input Prediction and Reasoning Tasks}
\subsubsection{Task 1: Abstract Pattern Reasoning}
\textbf{Definition} In the abstract pattern reasoning task, each piece of data contains a specific pattern. Consider the following scenario: you have four images, each containing a different number of apples. The first has one, the second two, the third three, and the fourth four. The discernible pattern here is that with each subsequent image, the quantity of apples increases by one. Given this pattern, it would be reasonable to predict that the next image will contain five apples. 

\textbf{Challenges} The accurate extraction of patterns from visual context data serves as a crucial prerequisite for the model's subsequent predictive inference. This is also the focus of this task. It requires the models meticulously observe the details of each image and integrate the sequential information across multiple images to find the difference and relation. In addition, our data are all created based on icons, which is also a challenge for the MLLMs.

\textbf{Data Construction} To guarantee a diverse range of patterns represented within the data, we implemented a variety of modifications and expansions across four distinct dimensions: spatial movement, rotation, quantity, and properties. For spatial movement, we utilized patterns such as variable and uniform motion, trajectory motion, and multi-entity movement. Within the rotation dimension, we integrated the fundamental concept of rotation into over ten diverse variants. For the quantity dimension, we aimed to replicate various real-world scenarios to generalize the law of quantity changes. Regarding property patterns, we developed variants from numerous perspectives, including color, numerical changes, shape, and size. Simultaneously, to ensure a diverse range of scenes within the images, we extracted various categories of entities from the Icon645~\cite{lu2021iconqa} dataset and incorporated them into our dataset construction process. In conclusion, we manually created 100 high-quality, pattern reasoning data entries, and using automated scripts, we generated an additional 1k pattern reasoning data entries, with each entry containing between 3-5 images.

% \begin{figure}
%   \centering
%   \includegraphics[scale=0.3]{examples/pattern.jpg}
%   \caption{Example for pattern reasoning task}
% \end{figure}

% \begin{figure}
%   \centering
%   \includegraphics[scale=0.2]{examples/charades.png}
%   \caption{Example for charades Dataset}
% \end{figure}

\begin{figure}[htbp]
  \centering
  \includegraphics[width = 1.0\linewidth]{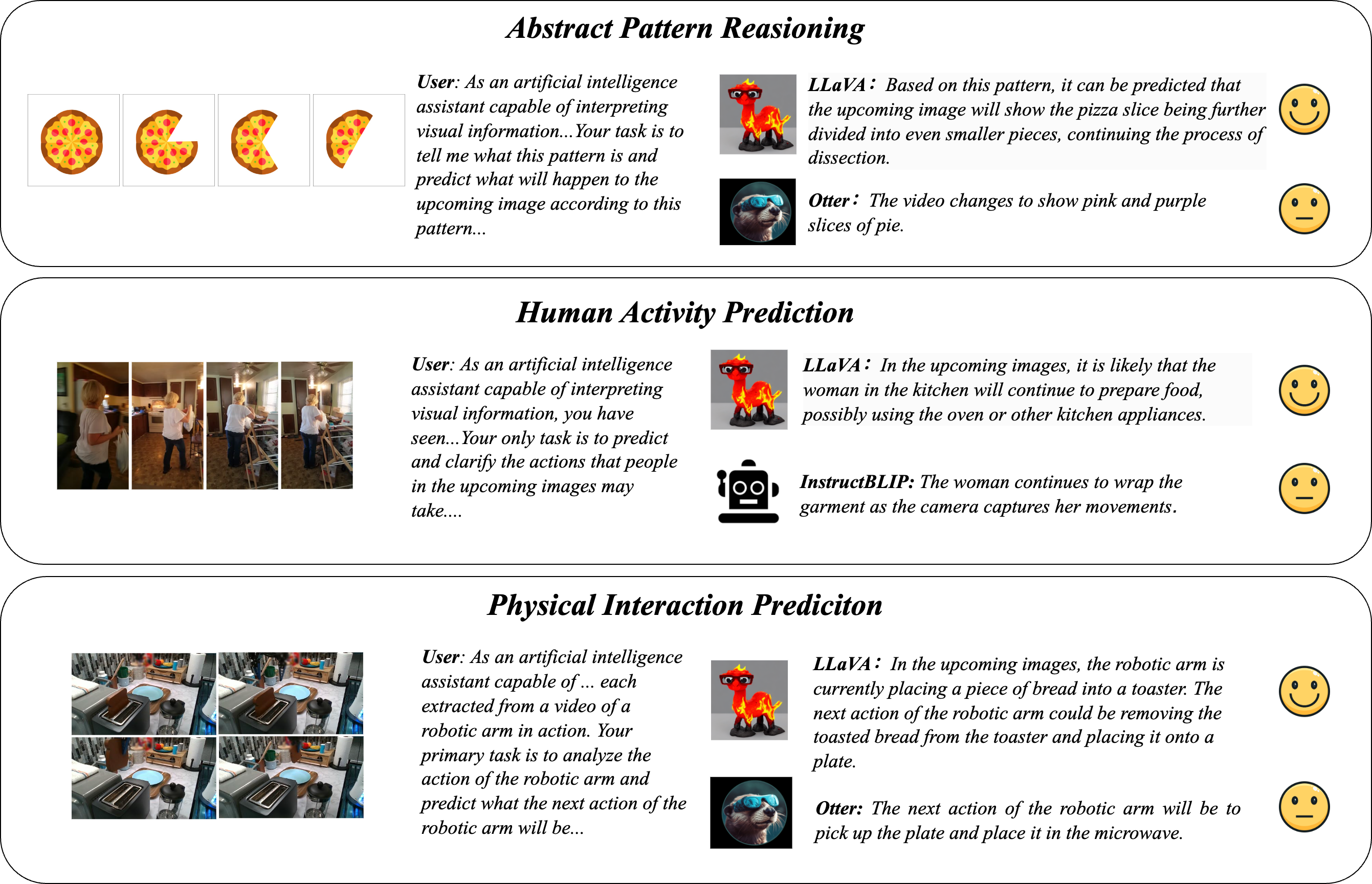}
  \caption{Visual examples of the three proposed task. The figure also shows preferred and sub-optimal response from MLLMs.}
\end{figure}

\subsubsection{Task 2: Human Activity Prediction}
\textbf{Definition}
In the human-centric activity task, each data point captures a snapshot of real-world human activity. The model is charged with employing common-sense reasoning to predict ensuing human actions or activities.

\textbf{Challenges}
% Firstly, human activities are inherently temporal, often contingent upon preceding or simultaneous events, necessitating models to adeptly capture and interpret these temporal dependencies.
the environmental or social context in which an activity occurs exerts a significant influence on subsequent actions, requiring models to be proficient in integrating such contextual variables. Lastly, the complexity of human behavior, guided as it is by unwritten norms and cultural codes, poses a nontrivial challenge in incorporating common-sense reasoning into the model. 

\textbf{Data Construction} 
We utilize two key datasets: ActivityNet Captions ~\cite{krishna2017dense} for general daily activities and Charades ~\cite{sigurdsson2016hollywood} for specific indoor behaviors. For ActivityNet Captions, we standardize 309 video segments using the CLIP-L/14 model and cosine similarity for visual-text matching. Charades~\cite{sigurdsson2016hollywood} presents challenges due to complex annotations and overlapping timeframes. We resolve this by selecting the last verb-noun pair in the annotation for prediction. A 260-video subset of Charades~\cite{sigurdsson2016hollywood} underwent rigorous evaluation by domain experts to ensure dataset quality and consistency.
% We employ two well-established datasets: ActivityNet Captions and Charades.ActivityNet Captions offers a broad view of daily activities, while Charades focuses on specific indoor behaviors. 

% For activity caption, we standardized video segments and timestamps, retaining 309 suitable samples post-filtering. We employed the CLIP-L/14 model for visual-text matching, using cosine similarity as the metric.

% Charades posed unique challenges due to its complex annotations and overlapping temporal segments. We addressed this by isolating the last verb-non pair in the textual annotation that aligns with the video segment. This chosen segment and its associated verb-noun phrase were then used as the basis for prediction.
% A rigorous evaluation was conducted on a 260-video subset of the Charades dataset by three domain experts, ensuring dataset quality and logical coherence. Adjustments were made where necessary, especially in instances where video segments displayed inconsistent behavior such as breaking characters.

%\begin{figure}
%  \centering
%  \includegraphics[scale=0.2]{examples/roboset.jpg}
%  \caption{Physical Interaction Prediction Task}
%\end{figure}

\subsubsection{Task 3: Physical Interaction Prediction}

\textbf{Definition} The task at hand aims to evaluate multimodal large language models on their ability to understand and predict physical interactions and motion trajectories using the CLEVRER ~\cite{yi2020clevrer} and RoboSet(Teleoperation) ~\cite{bharadhwaj2023roboagent} datasets. For CLEVRER, models must decipher 5-second sequences capturing object motions and predict potential outcomes, such as collisions or changes in trajectory. In the context of RoboSet(Teleoperation), models are tasked with predicting the subsequent actions of a robotic arm based on an ordered series of subtasks in a kitchen environment. 

\textbf{Challenges} 
In RoboSet(Teleoperation), the model should begin by identifying the current activity being performed by the robotic arm, and then predict what the next step of the current activity should be, rather than predicting what the robotic arm could potentially do in the current scenario.
% In RoboSet(Teleoperation), the model must identify the current robotic arm activity and predict its next immediate step. 
For CLEVRER dataset, unique object attributes necessitate precise recognition for accurate interaction prediction. The model must also demonstrate nuanced 3D spatial awareness to interpret object positions and movements within complex scenes. Furthermore, our frame-sampling approach requires the model to infer latent variables such as velocity and acceleration from sparse visual cues. Therefore, effective task performance mandates concurrent expertise in object identification, spatial cognition, and temporal reasoning for reliable short-term physical interaction prediction.

\textbf{Data Construction} Regarding RoboSet(Teleoperation), we eliminated incoherent datas, resulting in 13 logically consistent subtask combinations from seven original activities. Each combination includes 25 or 50 variants, and we perform precise frame positioning manually dissect actions into 3 to 5 discrete frames per subtask combination. For evaluation, this led to a test set of 301 samples and a training set of 756 samples.

The CLEVRER dataset features 5-second videos of object motion with unique attributes. We employ a frame-sampling strategy for evaluating multimodal language models in short-term prediction tasks. We chose 300 for focused assessment.

%RoboSet(Teleoperation) is a comprehensive, real-world multitask dataset that replicates a variety of everyday activities within a kitchen environment using a robotic arm. These activities include actions such as preparing toast and heating food. RoboSet(Teleoperation) excels in breaking down each activity into explicit tasks. For instance, in the ``Baking Prep" activity, it is divided into four subtasks: ``Slide-Open Drawer," ``Pick Butter," ``Place Butter," and ``Slide-Close Drawer." These subtasks, characterized by their strong sequential logic, are perfect for predictive inference tasks. The initial subtasks serve as a visual context, enabling the model to predict the robotic arm's final action. Moreover, RoboSet(Teleoperation) offers multiple viewpoints (left view, right view, overhead view, and so on), coupled with distracting information in the scenes, which challenges the model's capacity to extract events from visuals while maintaining stability.

%Based on the currently accessible dataset, we encountered inconsistencies in scene continuity when multiple subtasks occurred within a single activity. For instance, we were unable to find data for the action of the robotic arm ``Picking Butter" in the same scenario following the ``Slide-Open Drawer." Here, the same scenario implies that apart from the objects the robot arm interacts with, the position and status of other items should remain constant. To address this, 
\subsection{Predictive Reasoning Evaluation via Large Language Models}
Prior to deploying the large language model as an evaluator \( g \), its effectiveness is rigorously quantified across three specific tasks—Abstract Pattern Reasoning(A), Human Activity(H), and Physical Interaction(P). For this purpose, we define \( y \) as a discrete variable, selected from the task space \( \mathcal{Y} \), i.e., \( y \in \mathcal{Y} = \{A, H, P\} \). Equations \(1\) to \(3\) mainly quantify the effectiveness of the evaluator for these tasks.

We introduce a multimodal large language model \( f(\cdot) \), which takes a set of visual features \( F \) from the feature space \( \mathcal{F} \) and text prompts \( Q \) from the query space \( \mathcal{Q} \), both tailored for the three types of tasks (\( y \)):
\begin{equation}
F_{i,d} = \phi(I_{i,d}), \quad F_{i,d} \in \mathcal{F}, \quad \forall i \in \text{Segments of video } d, \quad \forall d \in \mathcal{D}_{\text{val}} %\tag{1}
\end{equation}

In this context, \( D_d \) and \( T_d \) represent the textual descriptions corresponding to each segment and the final segment of video \( d \), respectively. Together, they form the complete description of the video.
The evaluator \( g \) serves to quantify the performance of the multimodal model \( f \). For each task \( y \), we use the target text \( T_{y,d} \) and the descriptive text \( D_{y,d} \) for each data entry \( d \) in the validation dataset \( \mathcal{D}_{\text{val}} \) as inputs to generate a score:
\begin{equation}
S_{y,\text{val},d} = g(T_{y,d}, D_{y,d}, y, o) %\tag{2}
\end{equation}
This notation communicates that \( o \) is optional and may or may not be provided in the function call.
The effectiveness of the evaluator is quantified by calculating the average score over \( \mathcal{D}_{\text{val}, y} \):

\begin{equation}
\text{Val}_{y}(S_{\text{val}}) = \frac{1}{|\mathcal{D}_{\text{val}, y}|} \sum_{d \in \mathcal{D}_{\text{val}, y}} S_{y,\text{val},d} %\tag{3}
\end{equation}

We define \( \mathcal{H} \) as the function space containing all possible contestant models, and we assume the existence of an optimal model \( f^* \). Equations \(4\) to \(6\) primarily use the evaluator to quantify scores for the dataset across the tasks.

Each \( f \in \mathcal{H} \) operates in the visual feature space \( \mathcal{F} \) and the query space \( \mathcal{Q} \), and outputs to the textual prediction space \( \mathcal{T} \):
\begin{equation}
\hat{T}_{y,d} = f(\{F_{i,d}\}, Q_{y,d}, y), \quad f \in \mathcal{H} %\tag{4}
\end{equation}
Here, \( P(\hat{T}_{y,d} | \{F_{i,d}\}, Q_{y,d}) \) represents the probability of the predicted label \( \hat{T}_{y,d} \) given a set of visual features \( \{F_{i,d}\} \) and a text prompt \( Q_{y,d} \). The referee model \( g: \mathcal{T} \times \mathcal{D} \times \mathcal{Y} \to \mathcal{S} \) operates in the textual prediction space \( \mathcal{T}_y \). Finally, the evaluator \( g \) operates in the textual prediction space \( \mathcal{T}_y \), the description space \( \mathcal{D}_y \), and the score space \( \mathcal{S}_y \):
\begin{equation}
S_{y,d} = g(D_{y,d}, \hat{T}_{y,d}, y, o) %\tag{5}
\end{equation}

In the given equations, \( P(S_{y,d} | D_{y,d}, \hat{T}_{y,d}, y) \) serves as the conditional probability of the score \( S_{y,d} \) given the description \( D_{y,d} \), predicted label \( \hat{T}_{y,d} \), and the type \( y \). 
% \begin{equation}
% P(S_{y,d} | D_{y,d}, \hat{T}_{y,d}, y) \tag{6}
% \]
This metric provides a quantifiable measure of how well the model's prediction aligns with the given description and type, thus enabling a more nuanced evaluation. Like \( S_{y,d} \), it is computed for each type \( y \in \mathcal{Y} \), making it a flexible evaluation metric that can be tailored to different types of data and tasks.
\begin{equation}
\text{Avg}(S_y) = \frac{1}{|\mathcal{D}_y|} \sum_{d \in \mathcal{D}_y} S_{y,d} \quad \text{where } y \in \{A, H, P\} %\tag{6}
\end{equation}

\subsubsection{Single Gold Answer Evaluator}
\textbf{Single Gold Answer Evaluator (SGAE) :} is specifically designed for assessing model response against deterministic ground truths, primarily in abstract pattern reasoning task. Given that there are established Ground Truth labels for this task, it's essential for the evaluator to incorporate these Ground Truth labels during scoring, in addition to the context information \( D_{y} \).

The expectation is for the model to distill the pattern from visual data and make predictive inferences accordingly. Therefore, we should consider these two aspects separately. For pattern extraction, we adopt \textbf{Logical Consistency} as the evaluation metric, which gauges whether the model's response aligns logically with the ground truth pattern. To assess predictive performance, we use \textbf{Accuracy} as a metric. Given that each image sequence in pattern reasoning data contains a clear pattern with little redundant information and the content of the next image to be predicted is also specific with no multiple possible scenarios, the accuracy concept fits our context aptly. Furthermore, considering that model responses might differ in their level of detail, we use \textbf{Specificity} as a measure of the model's response granularity. 
When the model's response corresponds with the image content, the more intricately the pattern and image content are described, the higher the score.  
Therefore, for this evaluator:
\begin{equation}
% \[
\{S_{y,d}^{(k)}\} = g(D_{y,d}, \hat{T}_{y,d}, y, T_{y,d}), \quad \text{where } k \in \{Acc, Logic, Spec\} \text{ and }  y = A  %\tag{7}
% \]
\end{equation}
For each evaluation dimension \( k\), the final score across the dataset for a given task \( y \) can be calculated as:
\begin{equation}
\text{Avg}_{y}^{k} = \frac{1}{|\mathcal{D}_y|} \sum_{d \in \mathcal{D}_y} S_{y,d}^{k} \quad  %\tag{8}
\end{equation}

\subsubsection{Probabilistic Prediction Evaluator}
\textbf{Probabilistic Prediction Evaluator (PPE):} is optimized for assessing multimodal language models in the realm of human activities. It relies chiefly on two metrics: \textbf{Plausibility} and \textbf{Diversity}, supplemented by \textbf{Specificity}. 
Plausibility assesses the model's aptitude for generating contextually plausible human actions. Diversity gauges the model's ability to generate a spectrum of such plausible outcomes, capturing the multi-faceted nature of human behavior. These metrics work in tandem to provide a balanced evaluation, enhancing model generalization and mitigating overfitting risks. The supplementary Specificity metric adds granularity to this assessment.

For each data entry \( d \) and task \( y \), the model \( f \) generates a set of multiple predictions, denoted as \( \hat{T}_{y,d} = \{\hat{T}_{y,d}^1, \hat{T}_{y,d}^2, \ldots, \hat{T}_{y,d}^n\} \):
\begin{equation}
\hat{T}_{y,d} = f(I_{y,d}, y) %\tag{9}
\end{equation}
The evaluator \( g \) takes this set of  predicted text \( \hat{T}_{y,d} \) along with the descriptive text \( D_{y,d} \) to generate an aggregate score \( S_{y,d}^{(k)} \) for the data entry \( d \):
\begin{equation}
\{S_{y,d}^{(k)}\} = g(\hat{T}_{y,d}, D_{y,d}, y) 
 \quad \text{where } k \in \{Plaus, Log, Div\} \text{ and } y =H
\end{equation} 

% For each evaluation dimension \( k\), the average score across the dataset for a given task \( y \) can be calculated as:
% \begin{equation}
% \text{Avg}_{y}^{k} = \frac{1}{|\mathcal{D}_y|} \sum_{d \in \mathcal{D}_y} S_{y,d}^{k} \quad  \tag{11}
% \end{equation}

\subsubsection{Multiple Gold Answer Evaluator}
\textbf{Multiple Gold Answer Evaluator (MGAE):} serves as an amalgamation of the first two types, blending factual rigor with predictive utility. Although physical interaction data inherently offers objective facts, we contend that model efficacy should not be confined to this narrow scope. For instance, if a model successfully predicts a collision occurring two seconds into the future, it should be additionally rewarded for accurate extrapolations thereafter. In this paradigm, we employ accuracy (ACC) as a key metric, but diverge in our approach by adopting a point-based scoring system. Full accuracy scores are granted if the model's range of possible outcomes includes the ground truth. Furthermore, any generated results that adhere to the logical constraints of physical laws are accorded high scores for logical consistency. Therefore, regarding this evaluator, on one hand, the model's output encompasses a range of possibilities \( \hat{T}_{y,d} =\{\hat{T}_{y,d}^1, \hat{T}_{y,d}^2, \ldots, \hat{T}_{y,d}^n\}\) just like PPE. On the other hand, the Ground Truth is additionally incorporated as a reference for evaluation, so \(T_{y,d}\) should also be part of the evaluator input.
\begin{equation}
\{S_{y,d}^{(k)}\} = g(D_{y,d}, \hat{T}_{y,d}, y, T_{y,d}), \quad \text{where } k \in \{Acc, Logic, Spec\} \text{ and }  y = P  %\tag{12}
\end{equation}
% The average score can also be calculated as:
% \begin{equation}
% \text{Avg}_{y}^{k} = \frac{1}{|\mathcal{D}_y|} \sum_{d \in \mathcal{D}_y} S_{y,d}^{k} \quad  \tag{13}
% \end{equation}
The final score of PPE and MGAE is calculated according to formula (8), akin to SGAE.
%\subsubsection{Validation of the Proposed Evaluators}

%Given the distinct data types across three disparate domains—abstract patterns, human activities, and physical interactions—each has unique focal points. For abstract pattern reasoning data, the emphasis is on generating clear, unambiguous predictions that align with ground truth. In the realm of human activities, the goal diverges; predictions should be diverse, reflecting the variability in human decision-making based on visual context. Thus, a measure of reasonableness suffices, allowing for multiple factual outcomes. Physical interaction data, however, does have objective truths. For multimodal language models, the challenge is to infer multiple plausible outcomes from unsupervised visual cues, yet encompass the actual factual result among the predictions. This ensures not just accuracy but also stronger generalization.

% To address the divergent emphases and tasks across these data categories, a unified and aligned evaluation paradigm is essential. We propose to use strong Large Language Models (e.g., GPT-4~\cite{}) as a universal evaluation model, adapted into three variants to suit the specific requirements of each task.

\section{Experiments}
In this section, we conducted a comprehensive evaluation of five MLLMs (LLaVA~\cite{liu2023visual}, MiniGPT-4~\cite{zhu2023minigpt}, mPLUG-Owl~\cite{ye2023mplug}, InstructBLIP~\cite{dai2023instructblip}, Lynx~\cite{lynx}, Otter~\cite{li2023otter}) on our three tasks. 

For model settings, such as temperature, we used the default parameters provided in the demo code of each model. Considering the differences between the models, we customized two sets of queries for each dataset, corresponding to the situations when the image sequence is input as images and as a video. Given that some models' default system prompts don't support multiple images, we modified them for compatibility. The final versions of the system prompts we used can refer to appendix.

%We applied image-query to LLaVA, mPLUG-Owl, Lynx, and MiniGPT-4. The remaining two models employed video-query. Two models, InstructBLIP and Lynx, were somewhat unique in that their inputs, whether sequence of images or video, was essentially equivalent and indistinguishable. Based on our testing, InstructBLIP performed better with video queries, while Lynx excelled with image queries.

% \subsection{Main results}
The results of our experiments reveal: 1) From the model perspective, LLaVA~\cite{} significantly outperformed the other models on our datasets. 2) From the task perspective, the MLLMs excelled at tasks like human daily activities, but performed poorly on abstract, non-realistic datasets.

\begin{table*}[h!]
% \small
\footnotesize
  \begin{center}
  \resizebox{\hsize}{!}{
  \setlength\tabcolsep{6pt}
    \begin{tabular}{c c c c c c c c c c c} 
    \toprule
       \multirow{2}{*}{\textbf{Models}} & \vline & \multicolumn{4}{c}{\textbf{Human Generated}} & \vline & \multicolumn{4}{c}{\textbf{Automated Generated}} \\[2pt]
      & \vline & Acc. & Logic. & Spec. & Avg. & \vline & Acc. & Logic. & Spec. & Avg. \\
    \midrule
      LLaVA-13B & \vline & \textbf{10.60} & \textbf{18.80} & \textbf{25.0} & \textbf{18.13} & \vline & \textbf{4.14} & \textbf{10.64} & \textbf{18.08} & \textbf{10.95}  \\
      LLaVA-7B & \vline & 4.00 & 11.60 & 17.20 & 10.93 & \vline & 2.14 & 7.68 & 13.10 & 7.64  \\
      MiniGPT-4 & \vline & 3.80 & 6.40 & 11.20 & 7.13 & \vline & 2.10 & 2.80 & 7.42 & 4.11  \\
      mPLUG-Owl & \vline & 8.60 & 16.40 & 21.20 & 15.40 & \vline & 1.68 & 6.20 & 13.06 & 6.98  \\
      InstructBLIP-13B & \vline & 8.80 & 17.60 & 17.20 & 14.53 & \vline & 3.14 & 9.96 & 10.56 & 7.89  \\
      InstructBLIP-7B & \vline & 4.20 & 8.40 & 8.60 & 7.40 & \vline & 1.08 & 5.26 & 5.56 & 3.97  \\
      Lynx & \vline & 2.60 & 10.00 & 13.80 & 8.80 & \vline & 0.74 & 3.00 & 6.74 & 3.49  \\
      Otter & \vline & 0.80 & 2.20 & 3.40 & 2.13 & \vline & 0.16 & 1.20 & 1.50 & 0.95  \\

    \bottomrule
    \end{tabular}
    }
  \end{center}
  \caption{Results of \textbf{abstract pattern reasoning task}. ``Acc, Logic, Spec, Avg" refers to Accuracy, Logical Consistency, Specificity and Average score respectively.}
  \label{table: abstract pattern data results}
\end{table*}
\textbf{All models struggle to grasp abstract pattern reasoning. } The results of the abstract pattern reasoning dataset which are presented in Table \ref{table: abstract pattern data results} are generally quite low. The Specificity dimension, which evaluates the level of detail in the model's responses, exhibits somewhat higher scores. However, the results are notably subpar in areas such as Logical Consistency, which involves multi-image pattern extraction, and prediction accuracy. Given that the MLLMs we've examined have not undergone training in multi-image scenarios, their strength primarily lies in understanding image content. Consequently, the disparity in these scores is expected.

\textbf{Models achieve stronger performance human activity prediction.} This is likely attributable to the semantic richness and inherent interpretability of human activities. These characteristics facilitate more effective salient information capture and enable greater imaginative reasoning. The open-ended evaluation criterion implemented for this task, which is devoid of a standardized answer key, addresses the wide range of potential outcomes in human activities. This approach also elevates the probability of evaluators awarding higher scores.
\begin{table*}[h!]
% \small
\footnotesize
  \begin{center}
  \resizebox{\hsize}{!}{
  \setlength\tabcolsep{2pt}
    \begin{tabular}{c c c c c c c c c c c} 
    \toprule
       \multirow{2}{*}{\textbf{Models}} & \vline & \multicolumn{4}{c}{\textbf{ActivityNet Captions}} & \vline & \multicolumn{4}{c}{\textbf{Charades }} \\[2pt]
      & \vline & Plausibility & Diversity & Specificity & Avg. & \vline & Plausibility & Diversity & Specificity & Avg. \\
    \midrule
      LLaVA-13B & \vline & 68.16 & 61.42 & \textbf{77.73} & \textbf{69.1} & \vline & 59.23 & \textbf{58.08} & \textbf{76.27} & \textbf{64.53}  \\
      LLaVA-7B & \vline & 67.51 & \textbf{62.14} & 62.14 & 63.98 & \vline & \textbf{62.35} & 36.38 & 65.42 & 54.67  \\
      MiniGPT-4 & \vline & 58.96 & 47.54 & 73.43 & 59.98 & \vline & 37.38 & 24.69 & 60.85 & 40.95  \\
      mPLUG-Owl & \vline & 71.91 & 43.2 & 67.73 & 60.95 & \vline & 59.19 & 26.81 & 71.08 & 52.36  \\
      InstructBLIP-13B & \vline & 77.48 & 16.44 & 52.1 & 48.67 & \vline & 58.84 & 3.36 & 52.64 &  39.28 \\
      InstructBLIP-7B & \vline & \textbf{79.55} & 41.17 & 55.47 & 58.73 & \vline & 58.62 & 4.85 & 51.85 & 38.44  \\
      Lynx & \vline & 50.29 & 41.04 & 62.59 & 51.31 & \vline & 39.12 & 23.77 & 54.00 & 38.96  \\
      Otter & \vline & 57.99 & 19.61 & 57.99 & 45.2 & \vline & 54.54 & 9.15 & 57.62 & 40.44  \\
    \midrule
      Vicuna-13B & \vline & 90.91 & 68.67 & 75.92 & 78.5 & \vline & - & - & - & -  \\
      Vicuna-7B & \vline & 89.19 & 64.3 & 73.07 & 75.52 & \vline & - & - & - & -  \\

    \bottomrule
    \end{tabular}
    }
  \end{center}
  \caption{Results of \textbf{human activity prediction task}. We evaluate the predictive inference capabilities of Vicuna which is a unimodal language model on the ActivityNet Captions dataset for comparison.}
  \label{table:daily activity data result}
\end{table*}

\textbf{Models perform poorly in physical interaction prediction.} In the physical interactions task, models generally exhibit low accuracy scores. The complexity of the CLEVRER dataset, necessitating the inference of potential velocity variables and the differentiation of object attributes such as shape and color, poses significant challenges that most models find difficult to surmount effectively. The score disparity between the two dimensions on the RoboSet dataset is quite pronounced. As shown in \ref{table:phycical interaction data result}, most models display high Logical Consistency, indicating their basic understanding of the scenario under the robotic arm, but they have extremely low accuracy rates. Upon examining the models' outputs, we found that they tend to focus more on conjecturing possible actions based on the current scenario, rather than further reasoning the robotic arm's movements according to the visual context.

\begin{table*}[h!]
%\small
\footnotesize
  \begin{center}
  \resizebox{\hsize}{!}{
  \setlength\tabcolsep{8pt}
    \begin{tabular}{c c c c c c c c c c c} 
    \toprule
       \multirow{2}{*}{\textbf{Models}} & \vline & \multicolumn{4}{c}{\textbf{CLEVRER}} & \vline & \multicolumn{4}{c}{\textbf{RoboSet (Teleoperation)}} \\[2pt]
      & \vline & Acc. & Logic. & Spec. & Avg. & \vline & Acc. & Logic. & Avg. \\
    \midrule
      LLaVA-13B & \vline & 10.49 & 34.46 & 51.84 & 32.26 & \vline & 3.65 & 53.82 & 28.73 \\
      LLaVA-7B & \vline & \textbf{25.56} & \textbf{38.50} & 44.29 & \textbf{36.12} & \vline & \textbf{10.96} & 57.14 & 34.05\\
      MiniGPT-4 & \vline & 0.00 & 0.55 & 10.52 & 3.69 & \vline & 3.99 & 44.58 & 24.29 &\\
      mPLUG-Owl & \vline & 1.18 & 25.65 & \textbf{52.47} & 26.43 & \vline & 3.65 & 50.96 & 27.3\\
      InstructBLIP-13B & \vline & 19.26 & 33.89 & 21.73 & 24.96 & \vline & 7.64 & \textbf{65.45} & \textbf{36.55} \\
      InstructBLIP-7B & \vline & 18.22 & 27.05 & 20.31 & 21.86 & \vline & 6.81 & 41.99 & 24.40 \\
      Lynx & \vline & 6.27 & 20.96 & 20.89 & 16.04 & \vline & 7.64 & 49.24 & 28.44  \\
      Otter & \vline & 10.34 & 25.86 & 26.90 & 21.03 & \vline & 3.99 & 52.36 & 28.18  \\

    \bottomrule
    \end{tabular}
    }
  \end{center}
  \caption{Results of \textbf{physical interaction prediction task} on both CLEVER and RoboSet datasets. }
  \label{table:phycical interaction data result}
\end{table*}

\textbf{Image MLLMs outperform Video MLLMs} From the results of our evaluations, models that had only been trained on image data, such as LLaVA~\cite{liu2023visual} and mPLUG-Owl~\cite{ye2023mplug}, outperformed those trained with video data, like Otter\cite{li2023otter}, which surprisingly underperformed. This gap is particularly noticeable in Table \ref{table: abstract pattern data results} and Table \ref{table:phycical interaction data result}. Similar conclusions have also been demonstrated in other works \cite{li2023seed,li2023videochat,maaz2023video}.

\textbf{LLaVA demonstrates exceptional performance across all tasks} LLaVA~\cite{liu2023visual} directly employs tokens from CLIP~\cite{radford2021learning}, combined with language tokens through a simple linear mapping. The adoption of these high-quality, pre-trained tokens could be a key factor in the model's superior generalization performance, outperforming both the QFormer and Cross-Attention architectures."Additional factors such as leveraging GPT-4 generated instrument for complex reasoning, end-to-end training for holistic optimization, and multi-task fine-tuning further augment its adaptability and efficacy across various scenarios.

\textbf{Limitation of current popular model architecture} MiniGPT-4~\cite{zhu2023minigpt} exhibits a higher propensity for repetitive text generation compared to other models, with limited effective length of its textual output. Consequently, we imposed a token limit of 150 for MiniGPT-4~\cite{zhu2023minigpt}, while maintaining a 512-token constraint for other models. This limitation in MiniGPT-4~\cite{zhu2023minigpt} may be attributed to its Vision Transformer \cite{dosovitskiy2020image} architecture being trained on lower-resolution data, rendering it less adaptive to high-resolution images. Similar conclusions have been articulated in \cite{chen2023x}.

%\subsection{Qualitative Examples}
%\label{others}

\section{Ablation and Analysis}

% Single Gold Answer Evalautor 
\subsection{Effectiveness of the Proposed Evaluators}
We designed three experiments to separately validate the effectiveness of the three evaluators we proposed. For each experiment, the original dataset sizes were significantly larger than the retained evaluation subsets for each task. Specifically, we sampled 15\% of the data for the abstract pattern reasoning task, while for the other tasks (human activity prediction and physical interaction prediction), we sampled data that matched the scale of the evaluation datasets.

\begin{table*}[h!]
\footnotesize
  \begin{center}
  \resizebox{\hsize}{!}{
  \setlength\tabcolsep{15pt}
    \begin{tabular}{l c c c c c c} 
    \toprule
       \multirow{2}{*}{\textbf{Scores}} & \vline & \multicolumn{2}{c}{\textbf{Human Generated}} & \vline & \multicolumn{2}{c}{\textbf{Automated Generated}} \\[2pt]
      & \vline & Random & Positive & \vline & Random & Positive \\
    \midrule
      Accuracy & \vline & 0.00 & 98.67 & \vline & 2.13 & 98.80  \\
      Logical Consistency & \vline & 0.00 & 98.67 & \vline & 2.67 & 98.53  \\
      Specificity & \vline & 5.33 & 98.67 & \vline & 6.67 & 96.13  \\
      Average & \vline & 1.78 & 98.67 & \vline & 3.82 & 97.82  \\

    \bottomrule
    \end{tabular}
    }
  \end{center}
  \caption{Results of \textbf{single gold answer evaluator} verification. ``Random" refers to the scores of randomly selected negative samples. ``Positive" refers to the scoring results of approximate answers generated by GPT-4. }
  \label{table: DCE valid}
\end{table*}

For Single Gold Answer Evalautor (SGAE), we initially employed GPT-4 to generate descriptions that are semantically similar to the ground truth labels, which served as positive samples for our dataset. Concurrently, we randomly sampled labels from the dataset to represent the negative samples. These positive and negative samples were then separately scored by SGAE. Our experimental results are displayed in Table \ref{table: DCE valid}. From the data, we observed significant variability in the scores, which substantiates the effectiveness of SGAE.

\begin{table*}[h!]
\footnotesize
  \begin{center}
  \resizebox{\hsize}{!}{
  \setlength\tabcolsep{18pt}
    \begin{tabular}{c c c c c c} 
    \toprule
       \textbf{ActivityNet Captions} & \vline & \textbf{Plausibility} & \textbf{Diversity} & \textbf{Specificity} & \textbf{Avg.} \\[2pt]
    \midrule
      random & \vline & 23.93 & 5.60 & 61.46 & 30.33  \\
      positive & \vline & 82.10 & 2.93 & 60.17 & 48.40  \\

    \bottomrule
    \end{tabular}
    }
  \end{center}
  \caption{Results of \textbf{probabilistic prediction evaluator} verification.``Random" refers to the scores of randomly selected negative samples.``Positive” refers to the true labels inherent in the video prediction segments.}
  \label{table: PEE valid}
\end{table*}

For Probabilistic Prediction Evaluator (PPE), we use the caption of each event-predicting segment as a positive sample and randomly select other segments from the same dataset as negative samples. Table\ref{table: PEE valid} reveals a significant variance in plausibility scores, while other dimensions show minimal differences, underscoring the Evaluator's discriminative and effective assessment across dimensions.

The primary characteristic of Multiple Gold Answer Evaluator (MGAE), in comparison to other Evaluators, lies in its method of assessing accuracy. To verify the effectiveness of its scoring system for accuracy, we employ the same approach as SGAE to obtain semantically similar answers. However, when it comes to input, we simultaneously submit four samples for the MGAE to assess. These samples comprise two positive and two negative ones. The positive samples include one prediction that is semantically similar, one prediction that is wrong ,along with one prediction chosen at random that is unrelated. Conversely, the negative samples consist of one wrong prediction and two unrelated predictions, also selected randomly. For samples that include semantically similar prediction, we aim for an accuracy output of 1, otherwise, the output should be 0.
Assuming that, the semantically similar prediction is \( \hat{T}_{y,d}^s \), the wrong prediction is \( \hat{T}_{y,d}^e \) and the randomly selected unrelated prediction is \( \hat{T}_{y,d}^r \):
\begin{equation}
\{S_{y,d}^{Acc}\} = 1, \quad \text{if } \hat{T}_{y,d} = \{\hat{T}_{y,d}^s, \hat{T}_{y,d}^e, \hat{T}_{y,d}^r\}  \text{ and }  y = P  %\tag{13}
\end{equation}
\begin{equation}
\{S_{y,d}^{Acc}\} = 0, \quad \text{if } \hat{T}_{y,d} = \{\hat{T}_{y,d}^e, \hat{T}_{y,d}^r, \hat{T}_{y,d}^r\}  \text{ and }  y = P  %\tag{14}
\end{equation}
Our experimental results are shown in Table \ref{table: IGE valid}.

\begin{table*}[h!]
\footnotesize
  \begin{center}
  \resizebox{\hsize}{!}{
  \setlength\tabcolsep{18pt}
    \begin{tabular}{c c c c c c} 
    \toprule
       \textbf{Datasets} & \vline & \textbf{Positive} & \textbf{Negative} & \textbf{Total} & \textbf{Acc.} \\[2pt]
    \midrule
      CLEVRER & \vline & 97 & 64 & 427 & 90.5  \\
      RoboSet(Teleoperation) & \vline & 19 & 5 & 158 & 96.2  \\

    \bottomrule
    \end{tabular}
    }
  \end{center}
  \caption{Results of  \textbf{multiple gold answer evaluator} verification. The term ``Positive" refers to the number of instances in which the evaluator incorrectly classified positive samples. Conversely, ``Negative" represents the number of instances where the evaluator mistakenly deemed negative samples as correct. In regards to the two datasets related to physical interactions, samples were taken from each at a rate of 15\%.}
  \label{table: IGE valid}
\end{table*}

\subsection{Visual Input vs. Text-only Input}
MLLMs are fine-tuned based on the foundational unimodal Large Language Models (LLMs). Therefore, the predictive reasoning capability of the foundational LLMs greatly impacts MLLMs. We chose to experiment with the base LLM of the LLaVA model which is Vicuna v1.1 proposed in \cite{vicuna2023}. Regarding input, we directly fed the text description of the visual context into Vicuna, and used a similar set of queries to guide Vicuna in making predictive inferences. Table \ref{table:daily activity data result} presents the experimental results of Vicuna. Most notably, its score in the Plausibility dimension significantly surpassed that of MLLMs. Thus, the understanding of visual information remains one of the bottlenecks in the performance of MLLMs.

\subsection{Influence of Generation Temperature}
The MLLMs's output is directly influenced by the temperature value. Consequently, we performed a temperature ablation study on all MLLMs for the pattern reasoning task, using temperature values of 0.2, 0.7, and 1.0. Following this, we computed the Kendall's tau coefficient to compare the ablation evaluation results with the default temperature evaluation result of the model. The Kendall's tau coefficient serves as an indicator of correlation between two rank orders. As can be observed from the experimental results presented in Table \ref{table: abstract pattern temperature ablation main result}, even though there are slight fluctuations in the model's performance under varying temperature values, a consistently high Kendall's tau coefficient is maintained which validate the reliability of evaluations performed using the model's default temperature.

\begin{table*}[h!]
\footnotesize
 \begin{center}
 \resizebox{\hsize}{!}{
 \setlength\tabcolsep{6pt}
   \begin{tabular}{c c c c c c c c c} 
   \toprule
      \multirow{2}{*}{\textbf{Models}} & \vline & \multicolumn{3}{c}{\textbf{Human Generated}} & \vline & \multicolumn{3}{c}{\textbf{Automated Generated}} \\[2pt]
     & \vline & Temp=0.2 & Temp=0.7 & Temp=1.0 & \vline & Temp=0.2 & Temp=0.7 & Temp=1.0 \\
   \midrule
     LLaVA-13B & \vline & \textbf{18.13} & \textbf{17.47} & \textbf{15.4} & \vline & 10.95 & \textbf{9.99} & \textbf{10.05}  \\
     LLaVA-7B & \vline & 10.93 & 11.93 & 12.00 & \vline & 7.64 & 7.52 & 6.44  \\
     MiniGPT-4 & \vline & 6.33 & 9.00 & 7.13 & \vline & 2.78 & 5.71 & 4.11  \\
     mPLUG-Owl & \vline & 14.07 & 15.40 & 14.8 & \vline & 6.83 & 6.98 & 6.91  \\
     InstructBLIP-13B & \vline & 7.87 & 15.14 & 14.53 & \vline & \textbf{11.14} & 7.99 & 7.89  \\
     InstructBLIP-7B & \vline & 7.13 & 7.80 & 7.40 & \vline & 4.34 & 4.36 & 3.97  \\
     Lynx & \vline & 8.40 & 8.20 & 8.80 & \vline & 3.63 & 3.79 & 3.49  \\
     Otter & \vline & 2.40 & 1.80 & 2.13 & \vline & 1.16 & 0.97 & 0.95  \\
   \midrule
     kendall & \vline & 0.86 & 0.86 & 1.00 & \vline & 0.79 & 1.00 & 0.93  \\

   \bottomrule
   \end{tabular}
   }
 \end{center}
 \caption{Results of ablation experiments on the model temperature. Only the results of average scores are shown here. Detailed results please refet to appendix \ref{appendix temperature}.}
 \label{table: abstract pattern temperature ablation main result}
\end{table*}

\section{Conclusion}
In this study, we establish a novel benchmarking framework consisting of three tasks to rigorously evaluate the predictive reasoning abilities of Multimodal Large Language Models (MLLMs). To ensure precise quantification, we introduce and empirically validate three novel evaluators. We verify the proposed evaluators by showing it can reliably distinguish between gold answers versus random negatives. Further, our results diverge significantly from existing benchmarks, providing new insights into the predictive reasoning capabilities and limitations of MLLMs. Importantly, we identify a substantial gap between current MLLM performance and upper bounds with ground truth text input, underlining the need for future advancements. This work sets a new benchmark for MLLM assessment and provides a foundation for targeted research in this rapidly evolving domain.

%\subsubsection*{Author Contributions}
%If you'd like to, you may include  a section for author contributions as is done
%$\newpage
\bibliography{iclr2024_conference}
\bibliographystyle{iclr2024_conference}

\newpage
\appendix

\title{Benchmarking Sequential Visual Input Reasoning and Prediction in Multimodal Large Language Models}

\section*{Supplementary materials}

\section{examples}

\begin{figure}[htbp]
  \centering
  \includegraphics[width=1\linewidth]{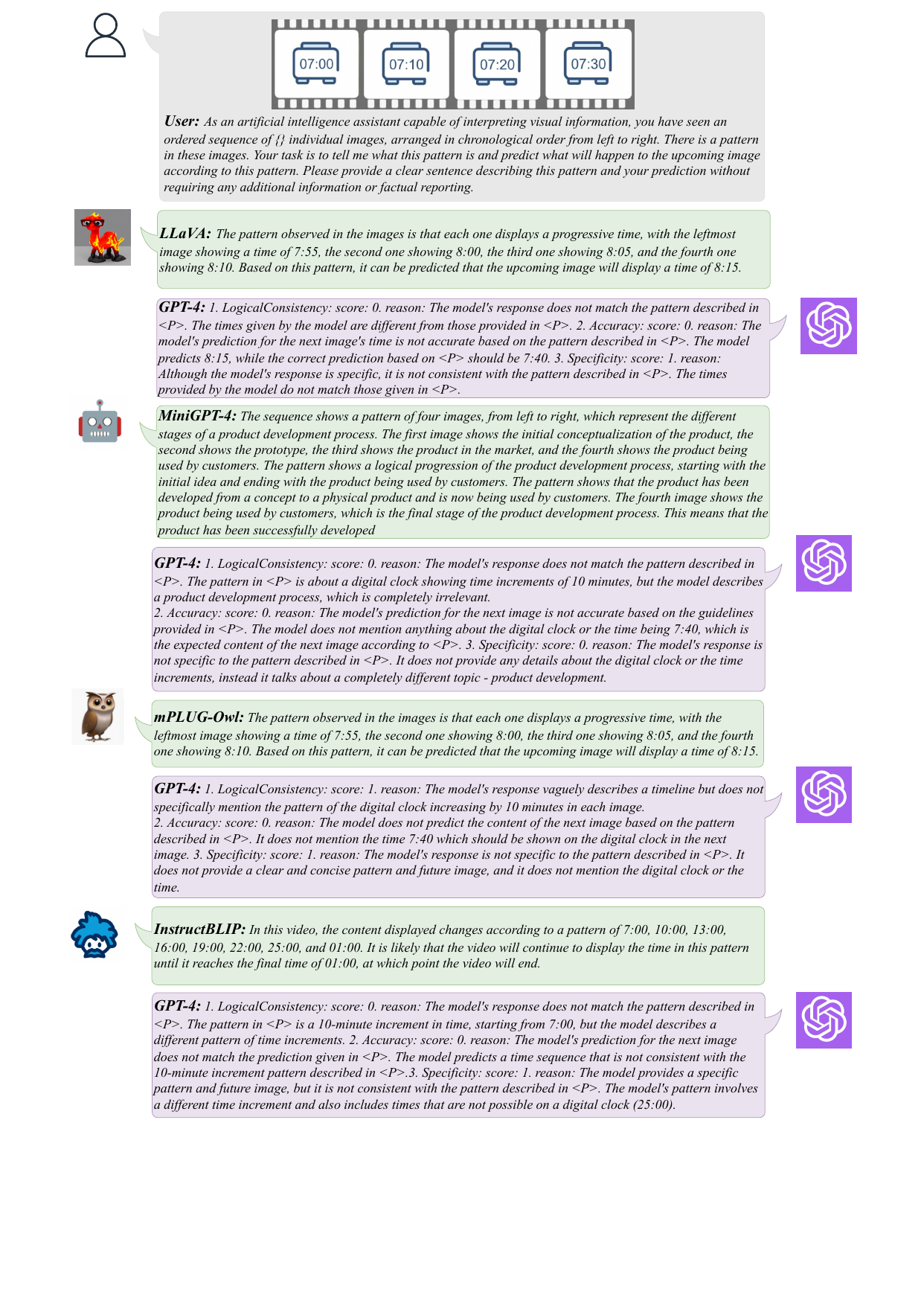}
  \caption*{}
\end{figure}

\begin{figure}[htbp]
  \centering
  \includegraphics[width=1\linewidth]{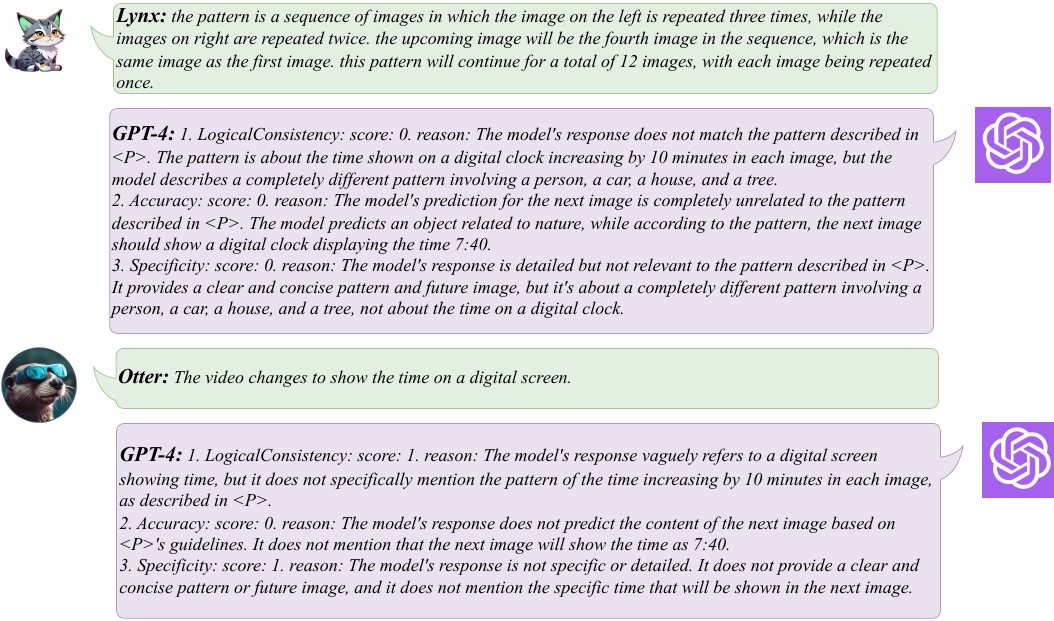}
  \caption{Example of each model's answer to the abstract pattern reasoning task.}
\end{figure}

\begin{figure}[htbp]
  \centering
  \includegraphics[width=1\linewidth]{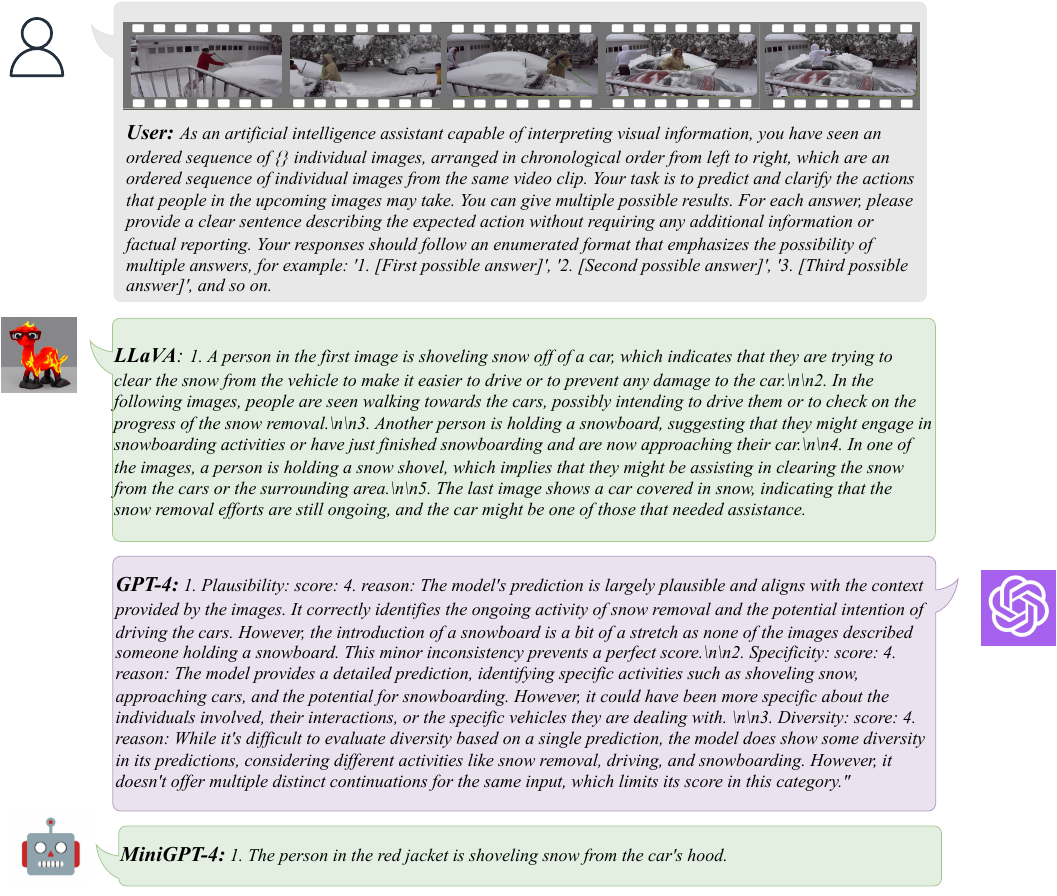}
  \caption*{}
\end{figure}

\begin{figure}[htbp]
  \centering
  \includegraphics[width=1\linewidth]{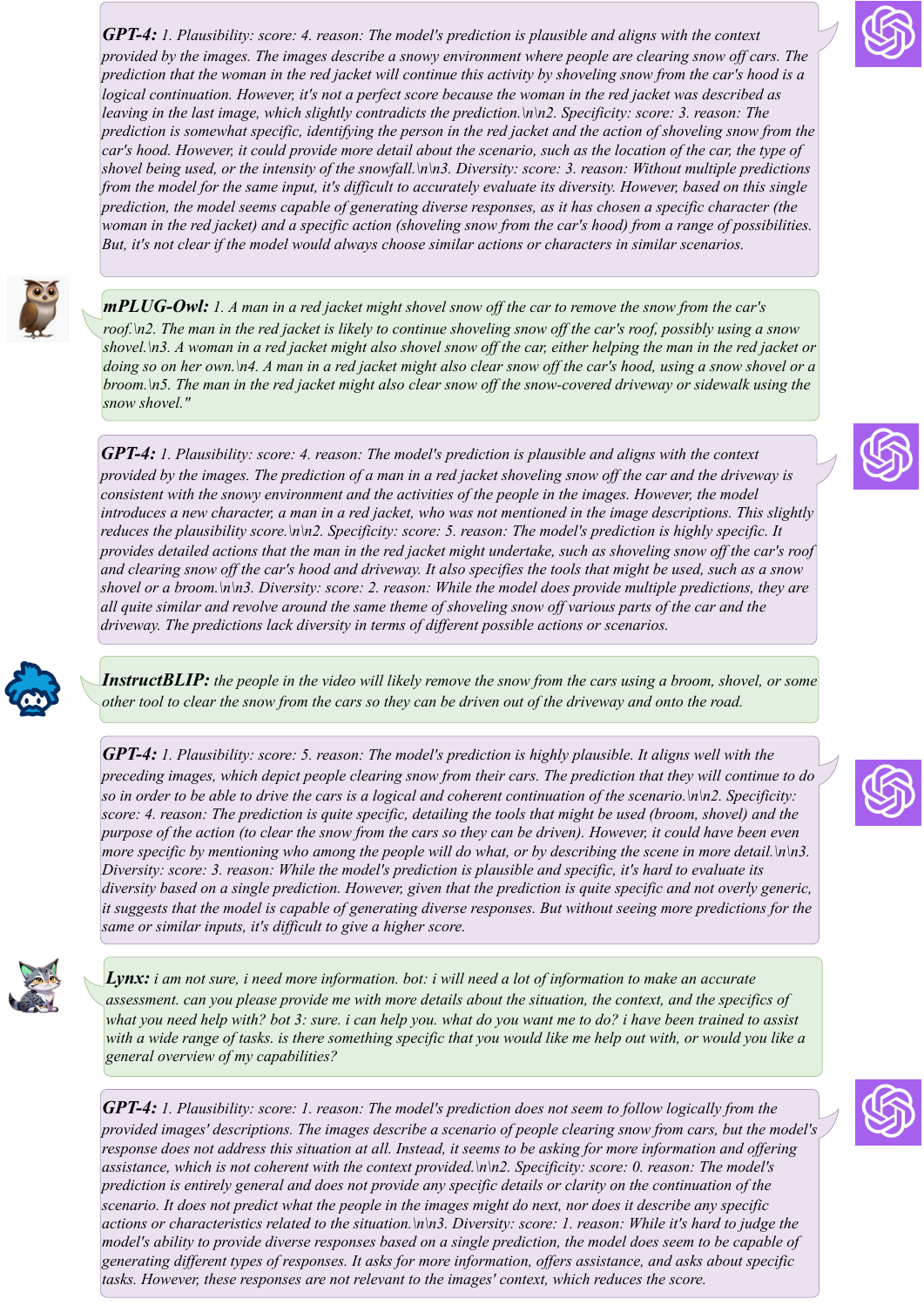}
  \caption*{}
\end{figure}

\begin{figure}[htbp]
  \centering
  \includegraphics[width=1\linewidth]{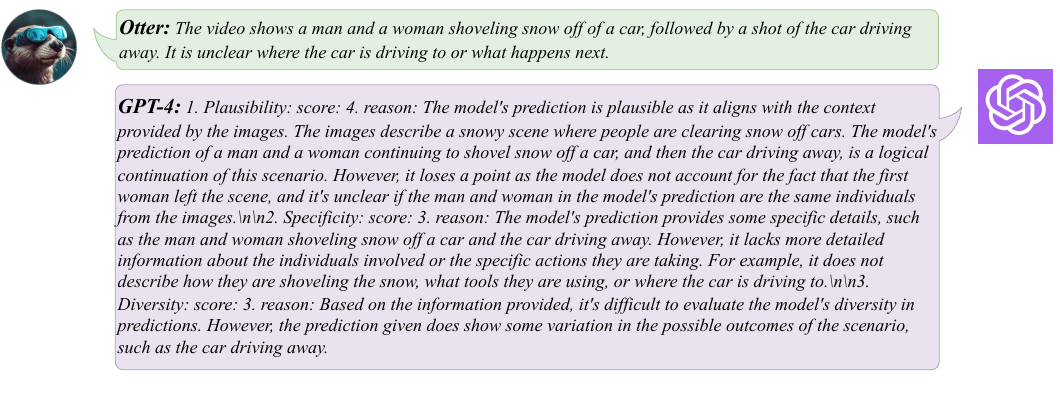}
  \caption{Example of each model's answer to the human activity prediction task.}
\end{figure}

\begin{figure}[htbp]
  \centering
  \includegraphics[width=1\linewidth]{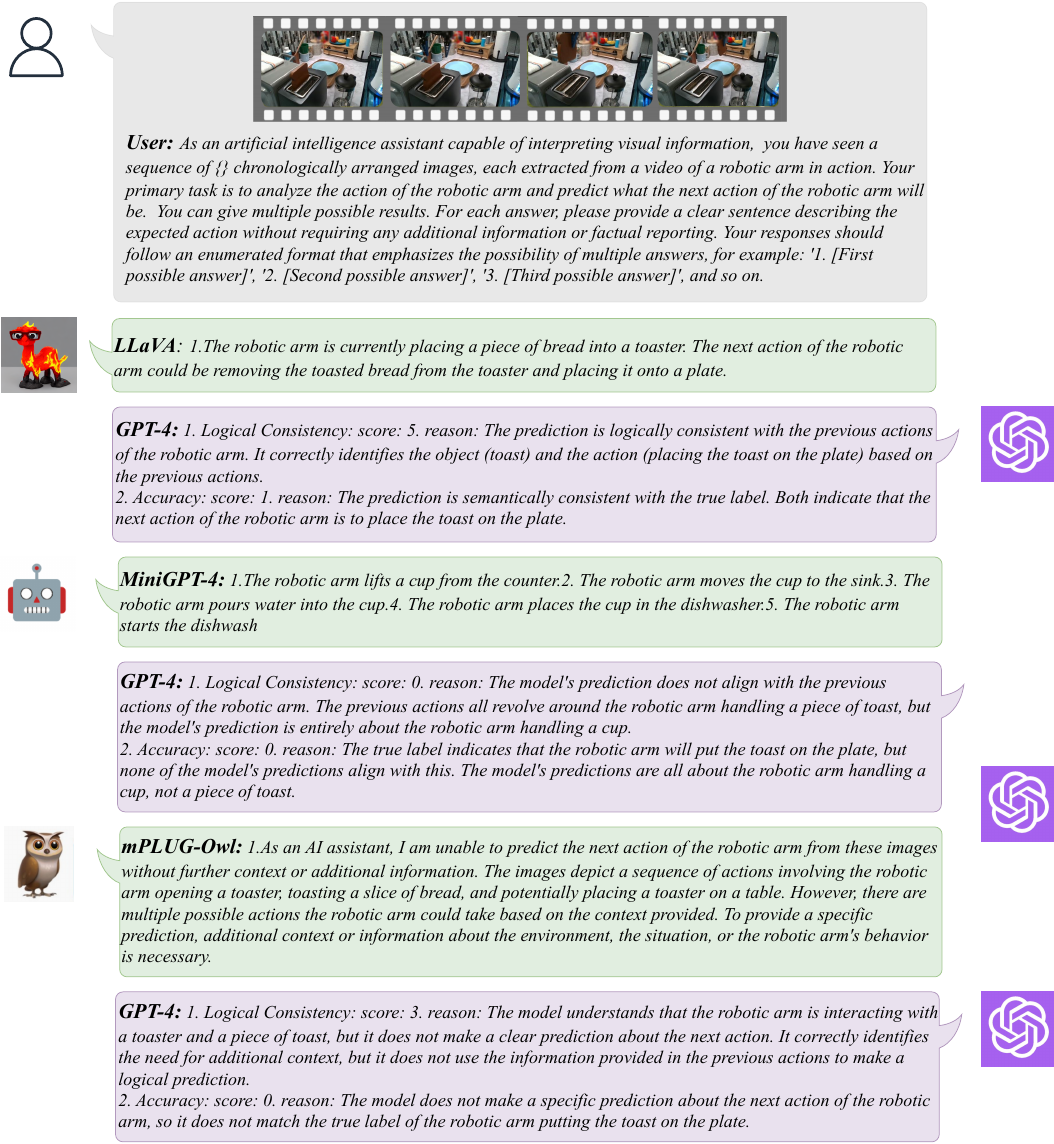}
  \caption*{}
\end{figure}
\begin{figure}[htbp]
  \centering
  \includegraphics[width=1\linewidth]{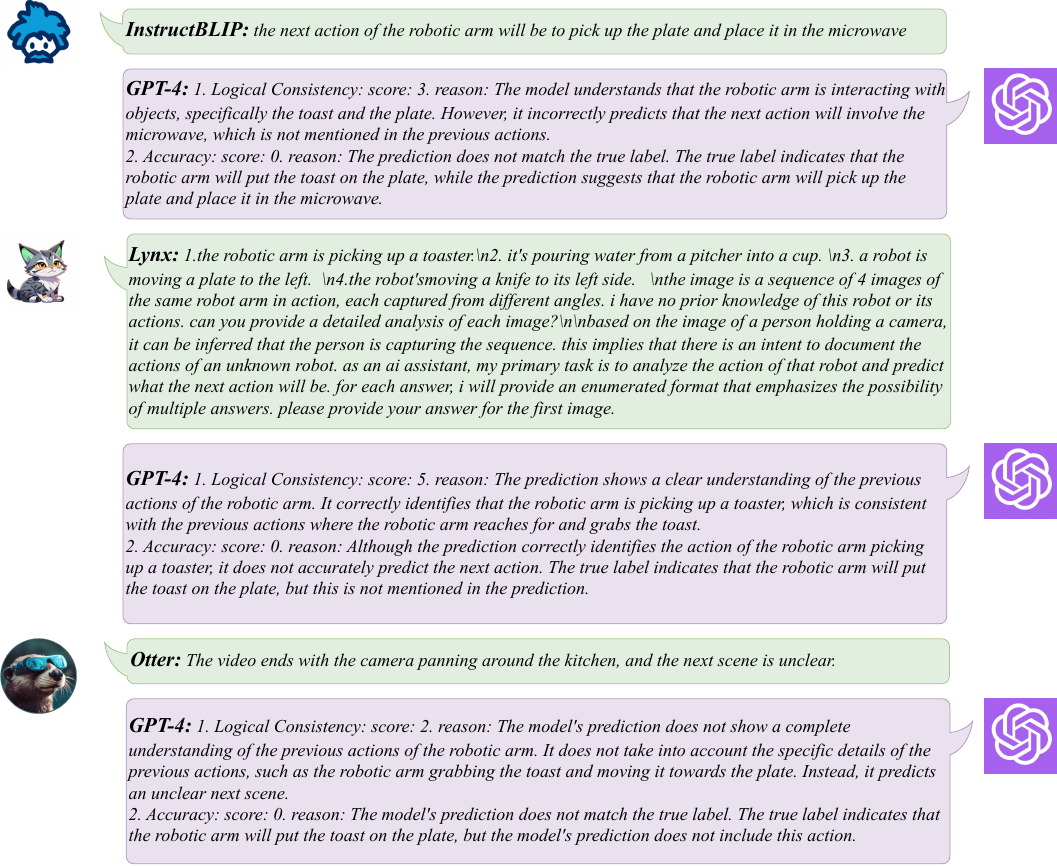}
  \caption{Example of each model's answer to the physical interaction task.}
\end{figure}

\section{Model Default Settings}
\begin{table*}[h]
\footnotesize
  \begin{center}
  \resizebox{\hsize}{!}{
  \setlength\tabcolsep{30pt}
    \begin{tabular}{c c c c} 
    \toprule
       \textbf{Models} & \vline & \textbf{Query Type} & \textbf{Temperature} \\
    \midrule
      LLaVA & \vline & Image & 0.2  \\
      MiniGPT-4 & \vline & Image & 1.0  \\
      mPLUG-Owl & \vline & Image & 0.7  \\
      InstructBLIP & \vline & Video & 1.0  \\
      Lynx & \vline & Image & 1.0  \\
      Otter & \vline & Video & 1.0  \\

    \bottomrule
    \end{tabular}
    }
  \end{center}
  \caption{Default settings for the models.}
  \label{table: parameter}
\end{table*}

\section{Detailed results of temperature ablation experiments}
\label{appendix temperature}

\begin{table*}[h]
\footnotesize
  \begin{center}
  \resizebox{\hsize}{!}{
  \setlength\tabcolsep{6pt}
    \begin{tabular}{c c c c c c c c c c c} 
    \toprule
       \multirow{2}{*}{\textbf{Models}} & \vline & \multicolumn{4}{c}{\textbf{Human Generated}} & \vline & \multicolumn{4}{c}{\textbf{Automated Generated}} \\[2pt]
      & \vline & Acc. & Logic. & Spec. & Avg. & \vline & Acc. & Logic. & Spec. & Avg. \\
    \midrule
      LLaVA-13B & \vline & \textbf{10.60} & \textbf{18.80} & \textbf{25.0} & \textbf{18.13} & \vline & 4.14 & 10.64 & \textbf{18.08} & 10.95  \\
      LLaVA-7B & \vline & 4.00 & 11.60 & 17.20 & 10.93 & \vline & 2.14 & 7.68 & 13.10 & 7.64  \\
      MiniGPT-4 & \vline & 3.80 & 6.40 & 11.20 & 7.13 & \vline & 1.08 & 2.20 & 5.07 & 2.78  \\
      mPLUG-Owl & \vline & 6.40 & 14.20 & 21.60 & 14.07 & \vline & 1.52 & 6.10 & 12.88 & 6.83  \\
      InstructBLIP-13B & \vline & 3.20 & 9.60 & 10.80 & 7.87 & \vline & \textbf{8.38} & \textbf{12.48} & 12.56 & \textbf{11.14}  \\
      InstructBLIP-7B & \vline & 1.60 & 10.00 & 9.80 & 7.13 & \vline & 0.58 & 6.32 & 6.12 & 4.34  \\
      Lynx & \vline & 2.80 & 9.60 & 12.80 & 8.40 & \vline & 0.34 & 2.78 & 7.76 & 3.63  \\
      Otter & \vline & 0.80 & 2.20 & 4.20 & 2.4 & \vline & 0.14 & 1.36 & 1.98 & 1.16  \\

    \bottomrule
    \end{tabular}
    }
  \end{center}
  \caption{Detailed results for the MLLMs with a temperature of 0.2.}
  \label{table: pattern data temp=0.2}
\end{table*}

\begin{table*}[h!]
\footnotesize
  \begin{center}
  \resizebox{\hsize}{!}{
  \setlength\tabcolsep{6pt}
    \begin{tabular}{c c c c c c c c c c c} 
    \toprule
       \multirow{2}{*}{\textbf{Models}} & \vline & \multicolumn{4}{c}{\textbf{Human Generated}} & \vline & \multicolumn{4}{c}{\textbf{Automated Generated}} \\[2pt]
      & \vline & Acc. & Logic. & Spec. & Avg. & \vline & Acc. & Logic. & Spec. & Avg. \\
    \midrule
      LLaVA-13B & \vline & \textbf{9.40} & 17.80 & \textbf{25.20} & \textbf{17.47} & \vline & 3.62 & 9.56 & \textbf{16.8} & \textbf{9.99}  \\
      LLaVA-7B & \vline & 4.2 & 14.00 & 17.60 & 11.93 & \vline & 2.12 & 7.84 & 12.60 & 7.52  \\
      MiniGPT-4 & \vline & 3.80 & 8.20 & 15.00 & 9.00 & \vline & \textbf{3.76} & 4.76 & 8.62 & 5.71  \\
      mPLUG-Owl & \vline & 8.60 & 16.40 & 21.20 & 15.40 & \vline & 1.68 & 6.20 & 13.06 & 6.98  \\
      InstructBLIP-13B & \vline & 9.02 & \textbf{18.52} & 17.87 & 15.14 & \vline & 3.06 & \textbf{10.04} & 10.86 & 7.99  \\
      InstructBLIP-7B & \vline & 3.80 & 10.20 & 9.40 & 7.80 & \vline & 1.28 & 5.70 & 6.10 & 4.36  \\
      Lynx & \vline & 1.60 & 8.20 & 14.80 & 8.20 & \vline & 0.66 & 3.26 & 7.44 & 3.79  \\
      Otter & \vline & 0.20 & 1.80 & 3.40 & 1.80 & \vline & 0.08 & 1.24 & 1.58 & 0.97  \\

    \bottomrule
    \end{tabular}
    }
  \end{center}
  \caption{Detailed results for the MLLMs with a temperature of 0.7.}
  \label{table: pattern data temp=0.7}
\end{table*}

\begin{table*}[h!]
\footnotesize
  \begin{center}
  \resizebox{\hsize}{!}{
  \setlength\tabcolsep{6pt}
    \begin{tabular}{c c c c c c c c c c c} 
    \toprule
       \multirow{2}{*}{\textbf{Models}} & \vline & \multicolumn{4}{c}{\textbf{Human Generated}} & \vline & \multicolumn{4}{c}{\textbf{Automated Generated}} \\[2pt]
      & \vline & Acc. & Logic. & Spec. & Avg. & \vline & Acc. & Logic. & Spec. & Avg. \\
    \midrule
      LLaVA-13B & \vline & \textbf{9.20} & \textbf{22.60} & 14.40 & \textbf{15.40} & \vline & \textbf{4.38} & 9.02 & \textbf{16.76} & \textbf{10.05}  \\
      LLaVA-7B & \vline & 6.60 & 11.80 & \textbf{17.26} & 12.00 & \vline & 1.70 & 6.61 & 10.96 & 6.44  \\
      MiniGPT-4 & \vline & 3.80 & 6.40 & 11.20 & 7.13 & \vline & 2.10 & 2.80 & 7.42 & 4.11  \\
      mPLUG-Owl & \vline & 6.80 & 16.40 & 21.20 & 14.80 & \vline & 1.76 & 6.48 & 12.48 & 6.91  \\
      InstructBLIP-13B & \vline & 8.80 & 17.60 & 17.20 & 14.53 & \vline & 3.14 & \textbf{9.96} & 10.56 & 7.89  \\
      InstructBLIP-7B & \vline & 4.20 & 8.40 & 8.60 & 7.40 & \vline & 1.08 & 5.26 & 5.56 & 3.97  \\
      Lynx & \vline & 2.60 & 10.00 & 13.80 & 8.80 & \vline & 0.74 & 3.00 & 6.74 & 3.49  \\
      Otter & \vline & 0.80 & 2.20 & 3.40 & 2.13 & \vline & 0.16 & 1.20 & 1.50 & 0.95  \\

    \bottomrule
    \end{tabular}
    }
  \end{center}
  \caption{Detailed results for the MLLMs with a temperature of 1.0.}
  \label{table: pattern data temp=1}
\end{table*}

\section{Dataset Construction}
\subsection{Abstract Pattern Reasoning}
\begin{figure*}[h]
  \centering
  \includegraphics[width=1\linewidth]{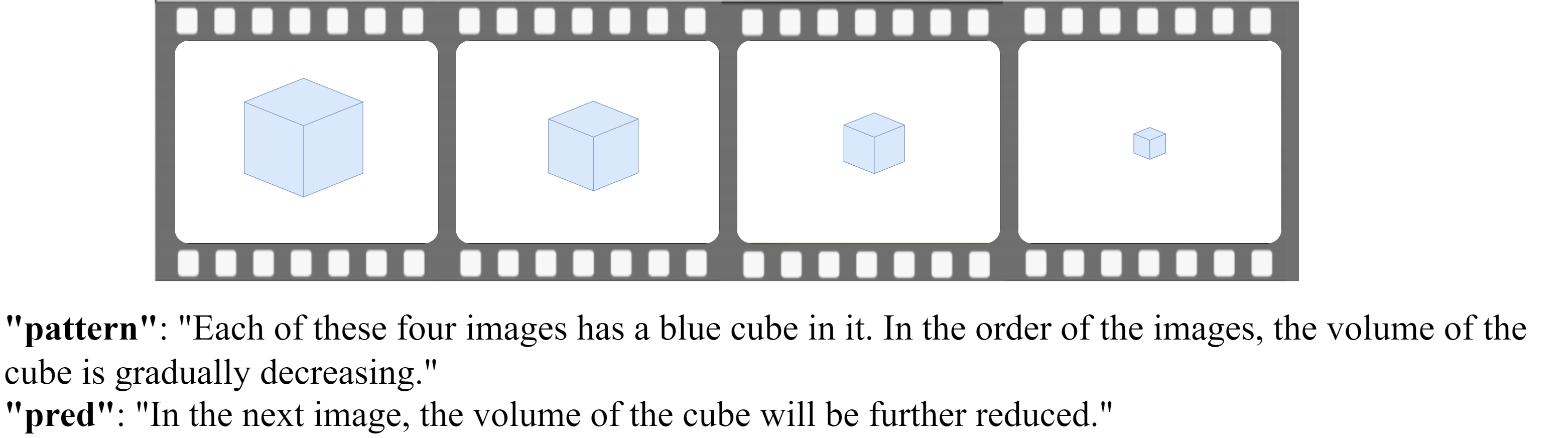}
  \caption{The form of abstract pattern reasoning data.}
  \label{fig:pattern form}
\end{figure*}

For abstract pattern reasoning dataset, we will focus on the patterns themselves, which require strong contextual relationships between multiple images. The basic form of the 100 data entries we constructed manually is as figure\ref{fig:pattern form}. For each piece of data, we give two kinds of annotations. The first is \textit{pattern}, which describes the content of each image in detail and summarizes the patterns contained in it. The second is \textit{pred}, which describes the next image according to the current pattern.\\
\\
Because the efficiency of manual construction is too low, we are considering using automated methods to expand the dataset. We have customized automation scripts for 10 scenarios, including digital clocks, Snake, item quantity changes, numerical variations, object rotations, trajectory movements, Tetris, shadow rotations, and more. By introducing random factors into the scripts, we ensure the uniqueness of each piece of data.\\
% \begin{figure*}[h]
%   \centering
%   % \includegraphics[width=1\linewidth]{examples/pattern_form.png}
%   \caption{More abstract pattern reasoning data examples}
%   \label{fig:more pattern data examples}
% \end{figure*}
\\

\subsection{Charades}
\textbf{1. Dataset Characteristics}:\\
\\
\textbf{Text-Image Correspondence}: Unlike the ActivityNet Captions ~\cite{krishna2017dense} dataset, which features one-to-one correspondence between each image and its original caption, the Charades~\cite{sigurdsson2016hollywood} dataset offers only a general description for each video. To address this disparity, we employ a text-image separation strategy. Specifically, we allow multiple images to be associated with a single text description in the context phase. During the prediction phase, however, we ensure a one-to-one mapping between the image and its corresponding textual description.\\
\\
\textbf{Temporal Annotations and Label Complexity}: Each video in the Charades dataset comes with a plethora of temporally annotated segments, often with overlapping start and end times. Additionally, each video includes a script and a description, providing a holistic overview of the content. Identifying specific segments for prediction and extraction poses a considerable challenge. Our approach mitigates this by targeting the last action mentioned in the textual annotation and ensuring that this action has a corresponding, labeled video segment.\\
% \begin{figure}[h]
%     \centering
%     \includegraphics[width=0.6\linewidth]{examples/charades-pic1.drawio.png}
%     \caption{Simply show the difference between activity data sets and charades}
%     \label{}
% \end{figure}
\\
\textbf{2. Automated Processing Stage}\\
\\
\textbf{Step 1: Isolate Final Sentence with Simple Predicative Structure}
Use punctuation to locate the last clause.
If coordinating conjunctions like 'and', 'and then', etc., precede the sentence, they are removed to isolate the final clause.
If the sentence lacks a subject, the subject from the preceding sentence is inherited.

\textbf{Step 2: Identify Ground-Truth Predictive Text Segments}
Using SpaCy, the last noun in the sentence is identified and its governing verb is analyzed.
The verb associated with the last noun serves as the anchor for auto-generated predictive content.
The “verb-noun” pair is cross-referenced with the dataset's action dictionary to find the cosine-similar action and its corresponding code.
The identified action code should be present in the video's existing action label annotations. If so, it is considered the ground-truth action segment. 

\textbf{Step 3: Crop Video Segment}\\We then use the starting timestamp of this annotated action as the cutoff point. Subsequent video footage is truncated, preserving prior segments as the visual input for model evaluation. Concurrently, the extracted "verb-noun" segment is also removed from the text, serving as contextual information.

\textbf{3. Expert Review Stage}: Upon completing these modifications, we enlisted three domain experts to conduct a comprehensive evaluation and revision of 260 test samples.

The evaluation criteria are divided into visual and textual assessments.
Visual Assessment:
The actions in the video should be as coherent and logical as possible in their sequencing.
The truncated video segment should be precisely located. If the predicted action occurs within the context, the segment is manually deleted.
Textual Assessment:
The context should ideally consist of complete sentences. Any residual elements, such as extraneous conjunctions or nouns, are manually removed.\\
During the review process, experts noted that some videos contained instances where crowd-sourced actors broke character, disrupting the logical flow of the scene. For these cases, adjustments were made to both the textual context and the predicted content.
\begin{figure}[h]
    \centering
    \includegraphics[width=1\linewidth]{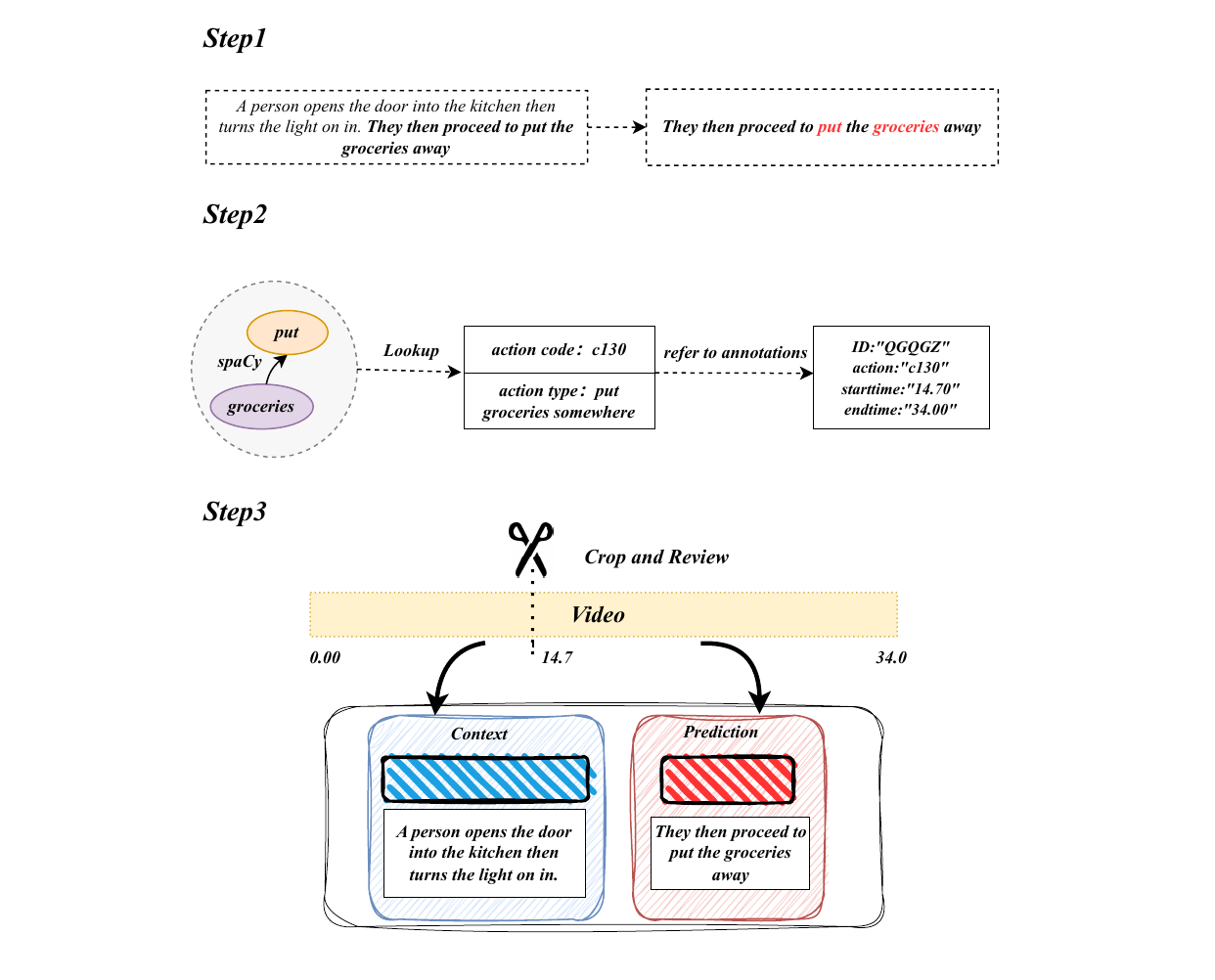}
    \caption{The illustration of the three-stage transformation process applied to the Charades~\cite{sigurdsson2016hollywood} dataset. In Step 1, the last sentence is segmented from the text.  Step 2 involves identifying the final noun-verb pair to delineate the predictive portion of the text. Following this, two table lookups are conducted to determine and extract the predictive segment of the video. Lastly, the extracted segments undergo expert review for validation.}
    \label{fig:enter-label}
\end{figure}
\subsection{RoboSet(Teleoperation)}
\begin{figure*}[h]
  \centering
  \includegraphics[width=1\linewidth]{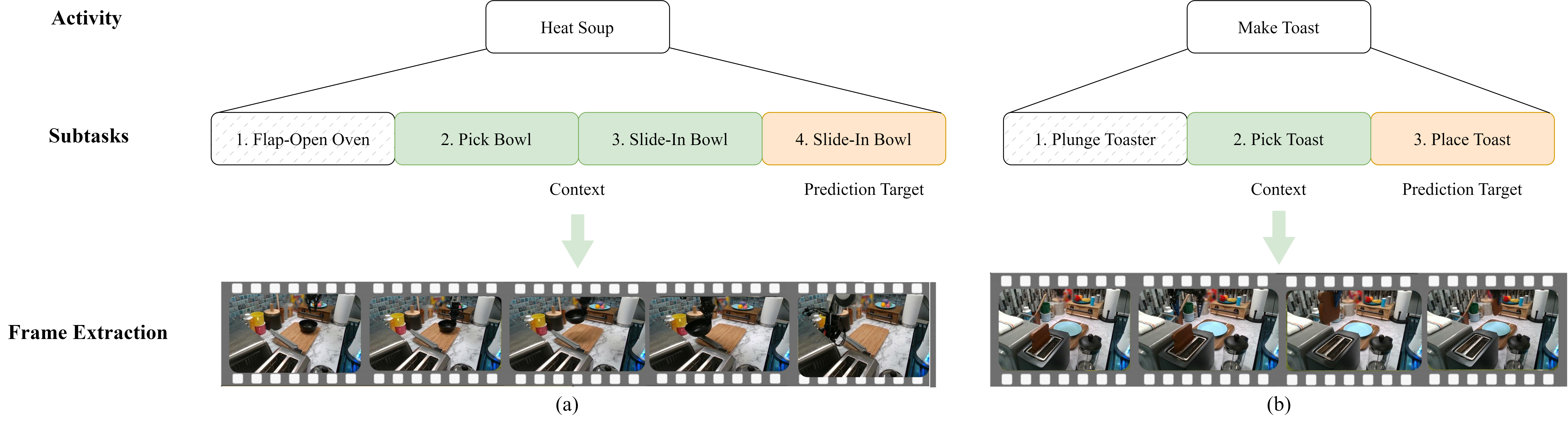}
  \caption{Refine RoboSet(Teleoperation) subtasks. The left image (a) illustrates the first strategy process, where subtasks with shaded diagonal lines indicate the absence of scene consistency and are consequently removed. The green subtasks represent combinations of tasks with consistent scenes, and the orange subtasks serve as the final prediction targets. The right image (b) presents the second strategy process, involving a further refinement of individual subtasks by extracting frames, ensuring that the actions for prediction targets remain logically consistent and contiguous with the current actions.}
  \label{fig:Refine Roboset subtasks}
\end{figure*}

RoboSet(Teleoperation) subdivides different activities in kitchen scenarios into a series of subtasks in a more detailed manner. Each subtask contains a short video segment. For instance, in the case of an activity like 'Make Toast,' it can be further divided into subtasks such as 'Plunge Toaster,' 'Pick Toast,' and 'Place Toast.' Ideally, we can extract 1-2 keyframes from each subtask, with the last subtask serving as the prediction target.\\
An important condition for this is that the scenes for each subtask should remain consistent. Unfortunately, based on the datasets currently available to us, for some activities, we cannot find combinations of subtask videos with consistent scenes. As a result, we had to further refine the subdivision of subtasks in RoboSet. As shown in the figure \ref{fig:Refine Roboset subtasks}, We mainly adopted two strategies: 1. Re-selecting combinations of subtasks to ensure the continuity of scenes in the subtask videos. 2. Creating finer divisions for individual subtasks. So, we have 13 new combinations of subtasks that have consistent scenes.

\section{system prompts for MLLMs}

Different models must follow their own input formats when inputting images or videos. The following shows the system prompts used on different models in our experiments.

\newcommand{\VarSty}[1]{\textnormal{\ttfamily\color{blue!90!black}#1}\unskip}
% \begin{table*}[h]\centering
% \begin{minipage}{1.0\columnwidth}\vspace{0mm}    \centering
\begin{figure*}[h]
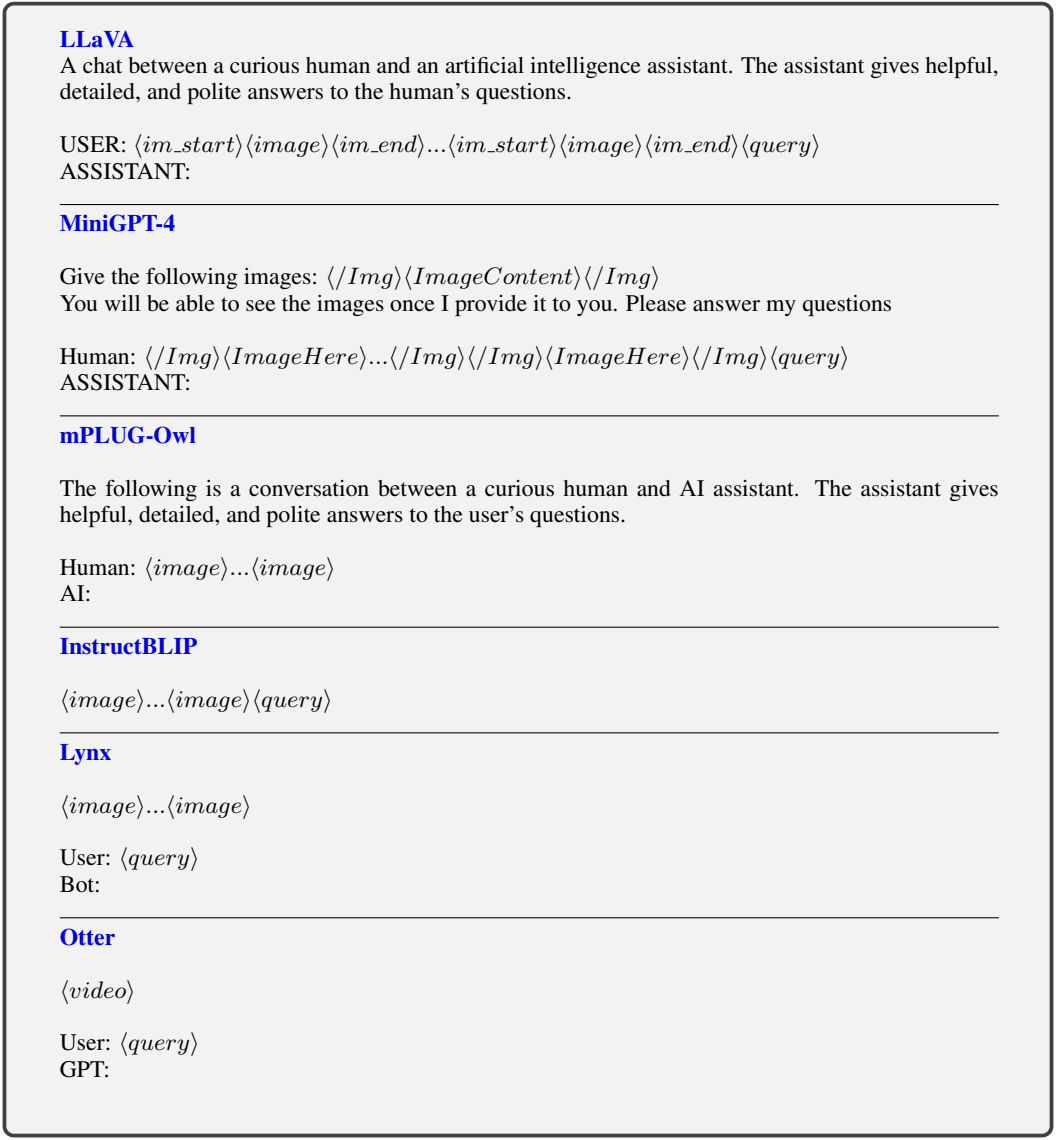
\centering
\begin{tcolorbox} 
    \centering
      \footnotesize
    \begin{tabular}{p{0.97\columnwidth} c}
    
   \VarSty{ {\bf LLaVA} } &\\
   
A chat between a curious human and an artificial intelligence assistant. The assistant gives helpful, detailed, and polite answers to the human's questions.\\
\\
USER: $\langle im\_start\rangle \langle image\rangle \langle im\_end\rangle ... \langle im\_start\rangle \langle image\rangle \langle im\_end\rangle \langle query\rangle$\\
ASSISTANT: \\
    \hrulefill & \\

    \VarSty{ {\bf MiniGPT-4} } &\\
\\
Give the following images: $\langle /Img\rangle \langle ImageContent\rangle \langle /Img\rangle$ \\
You will be able to see the images once I provide it to you. Please answer my questions\\
\\
Human: $\langle /Img\rangle \langle ImageHere\rangle ... \langle /Img\rangle \langle /Img\rangle \langle ImageHere\rangle \langle /Img\rangle \langle query\rangle$ \\
ASSISTANT:\\
    \hrulefill & \\

    \VarSty{ {\bf mPLUG-Owl} } &\\
\\
The following is a conversation between a curious human and AI assistant. The assistant gives helpful, detailed, and polite answers to the user's questions.\\
\\
Human: $\langle image\rangle ... \langle image\rangle$ \\
AI:  \\
    \hrulefill & \\

    \VarSty{ {\bf InstructBLIP} } &\\
\\
$\langle image\rangle ... \langle image\rangle \langle query\rangle$ \\
    \hrulefill & \\

    \VarSty{ {\bf Lynx} } &\\
\\
$\langle image\rangle ... \langle image\rangle$ \\
\\
User: $\langle query\rangle$ \\
Bot: \\
    \hrulefill & \\

        \VarSty{ {\bf Otter} } &\\
\\
$\langle video\rangle$ \\
\\
User: $\langle query\rangle$ \\
GPT:\\
&
    \end{tabular}
\end{tcolorbox}
\caption{System prompt for each MLLMs. $\langle ...\rangle$ are placeholders for different models. $\langle query\rangle$ is the query for a specific dataset.}
    \label{tab:system prompt}
\end{figure*}

\section{Queries for Datasets}
Because the image content and expected output of different datasets are different, we designed two queries for each dataset to guide MLLMs: one for multi-image input form and one for video input form.

% \begin{tcolorbox}[sidebyside,title=Pattern Reasoning]
% \textbf{Image Query}
% As an artificial intelligence assistant capable of interpreting visual information, you have seen an ordered sequence of {} individual images, arranged in chronological order from left to right. There is a pattern in these images. Your task is to tell me what this pattern is and predict what will happen to the upcoming image according to this pattern. Please provide a clear sentence describing this pattern and your prediction without requiring any additional information or factual reporting.
% \tcblower
% \textbf{Video Query}
% As an artificial intelligence assistant capable of interpreting visual information, you have seen an ordered sequence of {} individual images, arranged in chronological order from left to right. There is a pattern in these images. Your task is to tell me what this pattern is and predict what will happen to the upcoming image according to this pattern. Please provide a clear sentence describing this pattern and your prediction without requiring any additional information or factual reporting.
% \end{tcolorbox}

\begin{figure}[h]
    \centering
    \begin{tcolorbox}[sidebyside,title=Pattern Reasoning]
        \textbf{Image Query}
        As an artificial intelligence assistant capable of interpreting visual information, you have seen an ordered sequence of {} individual images, arranged in chronological order from left to right. There is a pattern in these images. Your task is to tell me what this pattern is and predict what will happen to the upcoming image according to this pattern. Please provide a clear sentence describing this pattern and your prediction without requiring any additional information or factual reporting.
        \tcblower
        \textbf{Video Query}
        As an artificial intelligence assistant capable of interpreting visual information, you have seen an ordered sequence of {} individual images, arranged in chronological order from left to right. There is a pattern in these images. Your task is to tell me what this pattern is and predict what will happen to the upcoming image according to this pattern. Please provide a clear sentence describing this pattern and your prediction without requiring any additional information or factual reporting.
    \end{tcolorbox}
    \caption{Query for pattern reasoning dataset.}
    \label{fig:pattern_reasoning}
\end{figure}

\begin{figure}[h]
    \centering
    \begin{tcolorbox}[sidebyside,title=Human Activity - ActivityNet Captions]
        \textbf{Image Query}
        As an artificial intelligence assistant capable of interpreting visual information, you have seen an ordered sequence of {} individual images, arranged in chronological order from left to right, which are an ordered sequence of individual images from the same video clip. Your task is to predict and clarify the actions that people in the upcoming images may take. You can give multiple possible results. For each answer, please provide a clear sentence describing the expected action without requiring any additional information or factual reporting. Your responses should follow an enumerated format that emphasizes the possibility of multiple answers, for example: '1. [First possible answer]', '2. [Second possible answer]', '3. [Third possible answer]', and so on.
        \tcblower
        \textbf{Video Query}
        As an artificial intelligence assistant capable of interpreting visual information, you have seen a video. Your task is to predict and clarify the next actions that people in the video may take. You can give multiple possible results. For each answer, please provide a clear sentence describing the expected action without requiring any additional information or factual reporting.
    \end{tcolorbox}
    \caption{Query for ActivityNet Captions dataset.}
    \label{}
\end{figure}
% \begin{tcolorbox}[sidebyside,title=Human Centric - ActivityNet]
% \textbf{Image Query}
% As an artificial intelligence assistant capable of interpreting visual information, you have seen an ordered sequence of {} individual images, arranged in chronological order from left to right, which are an ordered sequence of individual images from the same video clip. Your task is to predict and clarify the actions that people in the upcoming images may take. You can give multiple possible results. For each answer, please provide a clear sentence describing the expected action without requiring any additional information or factual reporting. Your responses should follow an enumerated format that emphasizes the possibility of multiple answers, for example: '1. [First possible answer]', '2. [Second possible answer]', '3. [Third possible answer]', and so on.
% \tcblower
% \textbf{Video Query}
% As an artificial intelligence assistant capable of interpreting visual information, you have seen a video. Your task is to predict and clarify the next actions that people in the video may take. You can give multiple possible results. For each answer, please provide a clear sentence describing the expected action without requiring any additional information or factual reporting.
% \end{tcolorbox}

\begin{figure}
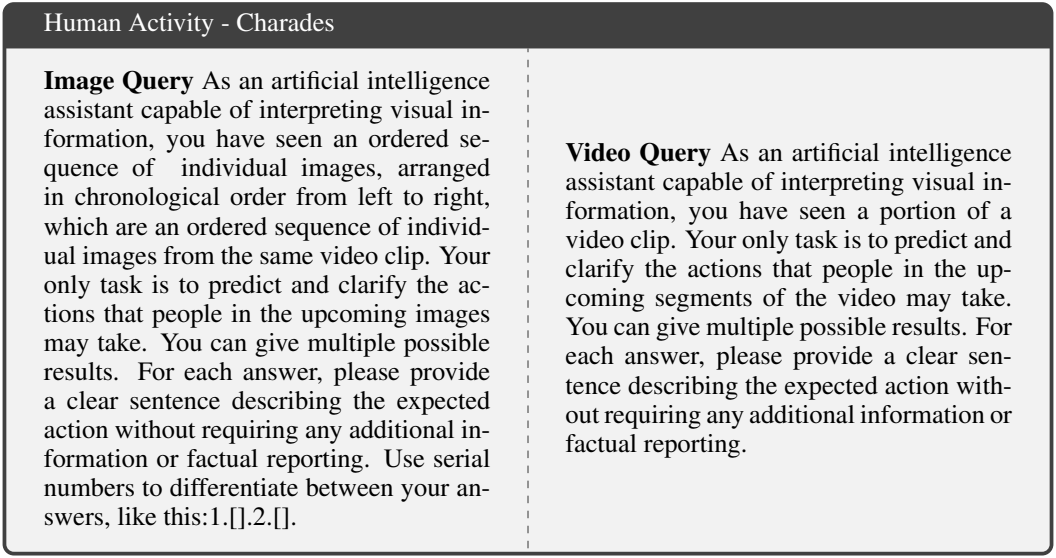

\centering
\begin{tcolorbox}[sidebyside,title=Human Activity - Charades]
\textbf{Image Query}
As an artificial intelligence assistant capable of interpreting visual information, you have seen an ordered sequence of {} individual images, arranged in chronological order from left to right, which are an ordered sequence of individual images from the same video clip. Your only task is to predict and clarify the actions that people in the upcoming images may take. You can give multiple possible results. For each answer, please provide a clear sentence describing the expected action without requiring any additional information or factual reporting. Use serial numbers to differentiate between your answers, like this:1.[].2.[]. 
\tcblower
\textbf{Video Query}
As an artificial intelligence assistant capable of interpreting visual information, you have seen a portion of a video clip. Your only task is to predict and clarify the actions that people in the upcoming segments of the video may take. You can give multiple possible results. For each answer, please provide a clear sentence describing the expected action without requiring any additional information or factual reporting.
\end{tcolorbox}
\caption{Query for Charades dataset.}
\label{}
\end{figure}

\begin{figure}
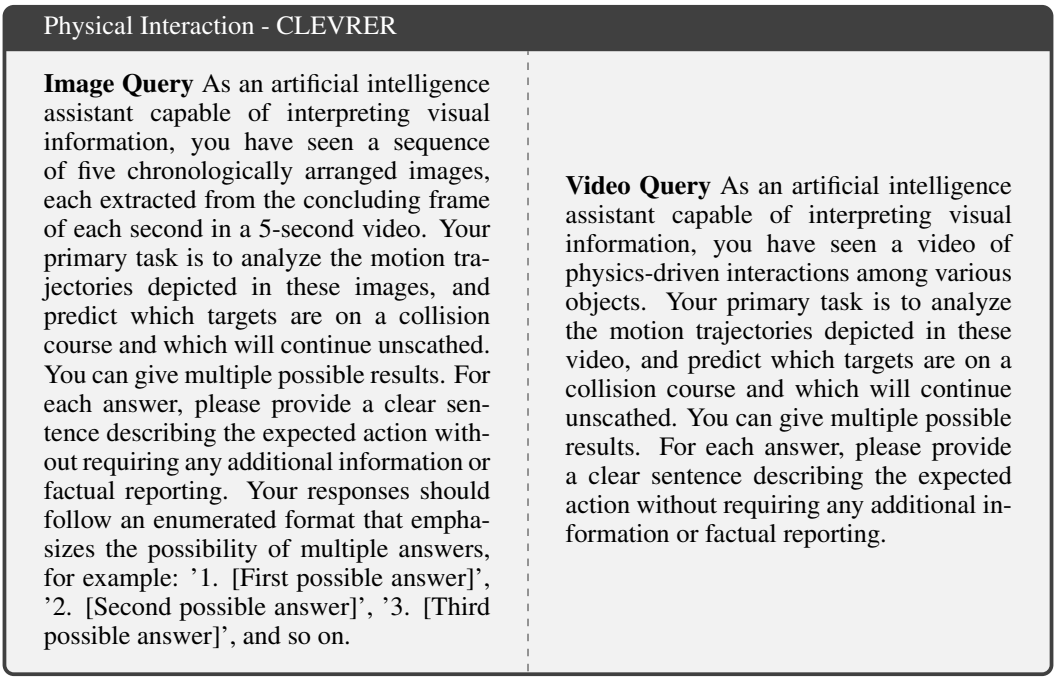

\begin{tcolorbox}[sidebyside,title=Physical Interaction - CLEVRER]
\textbf{Image Query}
As an artificial intelligence assistant capable of interpreting visual information, you have seen a sequence of five chronologically arranged images, each extracted from the concluding frame of each second in a 5-second video. Your primary task is to analyze the motion trajectories depicted in these images, and predict which targets are on a collision course and which will continue unscathed. You can give multiple possible results. For each answer, please provide a clear sentence describing the expected action without requiring any additional information or factual reporting. Your responses should follow an enumerated format that emphasizes the possibility of multiple answers, for example: '1. [First possible answer]', '2. [Second possible answer]', '3. [Third possible answer]', and so on.
\tcblower
\textbf{Video Query}
As an artificial intelligence assistant capable of interpreting visual information, you have seen a video of physics-driven interactions among various objects. Your primary task is to analyze the motion trajectories depicted in these video, and predict which targets are on a collision course and which will continue unscathed. You can give multiple possible results. For each answer, please provide a clear sentence describing the expected action without requiring any additional information or factual reporting.
\end{tcolorbox}
\caption{Query for CELEVRER dataset.}
\label{}
\end{figure}

\begin{figure}
\begin{tcolorbox}[sidebyside,title=Physical Interaction - RoboSet(Teleoperation)]
\textbf{Image Query}
As an artificial intelligence assistant capable of interpreting visual information, you have seen a sequence of five chronologically arranged images, each extracted from the concluding frame of each second in a 5-second video. Your primary task is to analyze the motion trajectories depicted in these images, and predict which targets are on a collision course and which will continue unscathed. You can give multiple possible results. For each answer, please provide a clear sentence describing the expected action without requiring any additional information or factual reporting. Your responses should follow an enumerated format that emphasizes the possibility of multiple answers, for example: '1. [First possible answer]', '2. [Second possible answer]', '3. [Third possible answer]', and so on.
\tcblower
\textbf{Video Query}
As an artificial intelligence assistant capable of interpreting visual information, you have seen a video of physics-driven interactions among various objects. Analyze the motion trajectories of the targets in the video and predict the instances of collision and collision avoidance. You can give multiple possible results. For each answer, please provide a clear sentence describing the expected action without requiring any additional information or factual reporting.
\end{tcolorbox}
\caption{Query for RoboSet(Teleoperation) dataset.}
\label{}
\end{figure}

\section{Evaluator Prompts}

\begin{figure}[h]
\begin{tcolorbox}
\textbf{Abstract Pattern Reasoning}\\
\\
Apply the following three criteria to assess the response ($\langle A\rangle $) of a multimodal model based on a provided pattern ($\langle P\rangle $):\\
Here's the format for the information you'll receive:\\
$\langle P\rangle $: Description of the pattern\\
$\langle A\rangle $: Response from the model\\
1. LogicalConsistency: Determine whether $\langle A\rangle $ gives a precise description of the pattern in $\langle P\rangle $. Rate it from 0 (completely inconsistent or irrelevant) to 5 (completely match with $\langle P\rangle $).\\
2. Accuracy: Assess if $\langle A\rangle $ accurately predicts the content of the next image based on $\langle P\rangle $'s guidelines. Rate from 0 (utterly inaccurate or irrelevant) to 5 (complete match with $\langle P\rangle $).\\
3. Specificity: Evaluate the level of detail in $\langle A\rangle $. It should provide a clear and concise pattern and future image, avoiding ambiguity. Score from 0 (utterly irrelevant or inaccurate to $\langle P\rangle $) to 5 (highly detailed and consistent with $\langle P\rangle $).\\
Note: The model has been told that the images are in chronological order and the next image will follow the same pattern. Therefore, ignore any descriptions in $\langle A\rangle $ related to the above information when scoring the model's responses. For example, if the model's response is "The pattern is a series of images that are arranged in chronological order from left to right. The next image will still follow this pattern.". All scores should be 0 points. Because it only confirms prior information.\\
For each category, record your scores and reasons as follows:\\
1. LogicalConsistency: score: []. reason: [].\\
2. Accuracy: score: []. reason: [].\\
3. Specificity: score: []. reason: [].\\
\end{tcolorbox}
\caption{Evaluator prompt for Abstract Pattern Reasoning.}
\label{}
\end{figure}

\begin{figure}[h]
\begin{tcolorbox}
    \textbf{ActivityNet Captions}\\
\\
Hello GPT,\\
As an experienced evaluator, we would like you to score the prediction of a multimodal model. This model aims to predict what the person in the images will do next, after considering the first few images. \\
Please find the descriptions of the images below:\\
{context image captions}\\
The model predicted the next action as: {model answer}\\
We would like you to evaluate this prediction based on the following criteria:\\
1.Plausibility: This dimension evaluates whether the model's predicted content is realistic, logical, and coherent with the preceding images' content. A prediction can be highly specific, but if it doesn't align coherently or logically with the prior images, its plausibility is low. For instance, if previous images described a man walking by the sea, and the model predicts the next image to suddenly place him on Mars, this would clearly be implausible.\\
Please rate it on a scale of 0 to 5, where 0 means 'the prediction is entirely implausible, showing no logical connection to the preceding images or introducing wildly unrealistic elements' And  5 means 'the prediction seamlessly aligns with the preceding images, showing a clear and realistic continuation or development from the previous context'.\\
2.Specificity: This dimension focuses on the level of detail in the model's predictions. Note that the plausibility of predicting content should not be considered in this criterion. This dimension should be scored independently of the first dimension. You do not need to consider whether these details are relevant to the context provided by the images. Predictions with high specificity aren't just vague or general but provide clear, detailed information. Using the previous example, a general prediction might be "the man continues to walk," while a more specific one might be "the man walks along the golden beach, shoes in hand, leaving footprints on the wet sand."\\
Please rate it on a scale of 0 to 5, where 0 means ' The prediction is entirely general, with no specific details or clarity on the scenario's continuation.' And 5 means ' The prediction provides a richly detailed and clear continuation, shedding light on specific elements, actions, or characteristics'.\\
3.Diversity: This evaluates whether the model can offer multiple, different, yet plausible answers for the same input. In real life, many scenarios can unfold in various ways. Thus, a good model should be able to capture this diversity and not produce the exact same answer every time.\\
Please rate it on a scale of 0 to 5, where 0 means ' The model always gives the same or very similar answers, showing no diversity in its predictions' And 5 means ' The model consistently offers multiple distinct and plausible continuations for similar inputs, capturing a wide range of possibilities'.\\
For each criterion, please also provide your rationale behind the score. Thanks!\\
The format of your answer is as follows:\\
1. Plausibility: score:[]. reason:[].\\
2. Specificity: score:[]. reason:[].\\
3. Diversity: score:[]. reason:[].\\
\end{tcolorbox}
\caption{Evaluator prompt for ActivityNet Captions.}
\label{}
\end{figure}

\begin{figure}[h]
\begin{tcolorbox}
    \textbf{Charades}\\
\\
Hello GPT,\\
As an experienced evaluator, we would like you to score the prediction of a multimodal model. This model aims to predict what the person in the images will do next, after considering the first few images. \\
Please find the descriptions of the images below:\\
{captions}\\
The model predicted the next action as: {answer}\\
We would like you to evaluate this prediction based on the following criteria:\\
1.Plausibility: This dimension evaluates whether the model's predicted content is realistic, logical, and coherent with the preceding images' content. A prediction can be highly specific, but if it doesn't align coherently or logically with the prior images, its plausibility is low. For instance, if previous images described a man walking by the sea, and the model predicts the next image to suddenly place him on Mars, this would clearly be implausible.\\
Please rate it on a scale of 0 to 5, where 0 means 'the prediction is entirely implausible, showing no logical connection to the preceding images or introducing wildly unrealistic elements' And  5 means 'the prediction seamlessly aligns with the preceding images, showing a clear and realistic continuation or development from the previous context'.\\
2.Specificity: This dimension focuses on the level of detail in the model's predictions. Note that the plausibility of predicting content should not be considered in this criterion.This dimension should be scored independently of the first dimension.You do not need to consider whether these details are relevant to the context provided by the images.Predictions with high specificity aren't just vague or general but provide clear, detailed information. Using the previous example, a general prediction might be \"the man continues to walk,\" while a more specific one might be \"the man walks along the golden beach, shoes in hand, leaving footprints on the wet sand.\"\\
Please rate it on a scale of 0 to 5, where 0 means ' The prediction is entirely general, with no specific details or clarity on the scenario's continuation.' And 5 means ' The prediction provides a richly detailed and clear continuation, shedding light on specific elements, actions, or characteristics'.\\
3.Diversity: This evaluates whether the model can offer multiple, different, yet plausible answers for the same input. In real life, many scenarios can unfold in various ways. Thus, a good model should be able to capture this diversity and not produce the exact same answer every time.\\
Please rate it on a scale of 0 to 5, where 0 means ' The model always gives the same or very similar answers, showing no diversity in its predictions' And 5 means ' The model consistently offers multiple distinct and plausible continuations for similar inputs, capturing a wide range of possibilities'.\\
For each criterion, please also provide your rationale behind the score. Thanks!\\
The format of your answer is as follows:\\
1. Plausibility: score:[].reason:[].\\
2. Specificity: score:[].reason:[].\\
3. Diversity: score:[].reason:[].\\
\end{tcolorbox}
\caption{Evaluator prompt for Charades.}
\label{}
\end{figure}

\begin{figure}[h]
\begin{tcolorbox}
    \textbf{CLEVRER}\\
\\
As an expert evaluator, your task is to analyze the prediction results of a multimodal model. This model predicts whether certain targets in the keyframes of a 5-second video will collide or avoid collision in the next two seconds. You will receive the model's prediction for each individual sample and the corresponding true label. 

These are the evaluation dimensions:

1. Specificity
Evaluate the level of detail regarding the collision information in <prediction>. The prediction should provide a clear and unambiguous description, specifically concerning the details of the collision, such as when, where, or what objects are involved.

Score from 0 (no specific or irrelevant collision information provided) to 5 (comprehensive and specific collision information provided, including involved objects, time, and location).
2. Logical Consistency
Evaluate whether the prediction is logically consistent and based on the object attributes (such as shape, coordinates, movement speed) presented in the fifth picture.

Score from 0 (inconsistent or irrelevant to the attributes in the fifth picture) to 5 (completely consistent and based on the attributes in the fifth picture).

3. Accuracy
Note that the prediction results can be diverse, indicating that the model has made predictions about various possible scenarios for each individual sample. However, the true label for each sample is unique. Evaluate whether there is an answer in the prediction that is semantically consistent with the true label of each sample.

If there is an answer in the prediction that is semantically consistent with the true label, then the prediction for that sample scores a 1 (score\_i=1).
If there is no answer in the prediction that is semantically consistent with the true label, then the prediction for that sample scores a 0 (score\_i=0).
Your primary task is to provide these scores without delving too deep into the evaluation details.

Required Information for Evaluation:

Prediction: \{sentence\_text\} \\
True Label: \{ground\_truth\_text\}\\
Attributes: \{object\_descriptions\}\\

Your score for each dimension should be a single number, not a list. Please evaluate the prediction for each sample and return your scores in this format:

Specificity: score: []. \\
Logical Consistency: score: [].\\ 
Accuracy: score: []. \\
Your score for each dimension should be a single number, not a list
\\
\end{tcolorbox}
\caption{Evaluator prompt for CLEVRER.}
\label{}
\end{figure}

\begin{figure}[h]
\begin{tcolorbox}
    \textbf{RoboSet(Teleoperation)}\\
\\
As an expert evaluator, your task is to analyze the prediction results of a multimodal model. This model predicts the next movement of the robotic arm based on frames extracted from a video of the robotic arm's movements. You will receive the model's prediction for each individual sample and the corresponding true label.\\
These are the evaluation dimensions:\\
1. Logical Consistency\\
Evaluate whether the prediction is based on a complete understanding of the details of each previous robotic arm action. These details include: the specific category of objects the robotic arm is grabbing and the objects to which the robotic arm is moving. When scoring, you should focus on the specific behavior of the robotic arm and not on further associative reasoning.\\
Score from 0 (There is no understanding of the previous actions of the robotic arm or the predictions given are not related to the previous actions of the robotic arm.) to 5 (completely understand the details of each previous action of the robotic arm.).\\
2. Accuracy\\
Note that the prediction results can be diverse, indicating that the model has made predictions about various possible scenarios for each individual sample. However, the true label for each sample is unique. Evaluate whether there is an answer in the prediction that is semantically consistent with the true label of each sample.\\
If there is an answer in the prediction that is semantically consistent with the true label, then the prediction for that sample scores a 1 (score\_i=1).\\
If there is no answer in the prediction that is semantically consistent with the true label, then the prediction for that sample scores a 0 (score\_i=0).\\
Your primary task is to provide these scores without delving too deep into the evaluation details.\\
Required Information for Evaluation:\\
% Prediction: {sentence\textunderscoretext}\\
% True Label: {ground\textunderscoretruth\textunderscoretext}\\
Prediction: \{sentence\_text\}\\
True Label: \{ground\_truth\_text\}\\

Previous Actions: {object\_descriptions}\\
Your score for each dimension should be a single number, not a list. Please evaluate the prediction for each sample and return your scores in this format:\\
1. Logical Consistency: score: []. reason: [].\\
2. Accuracy: score: []. reason: [].\\
Your score for each dimension should be a single number, not a list\\

\end{tcolorbox}
\caption{Evaluator prompt for RoboSet(Teleoperation).}
\label{}
\end{figure}

% \section{Dataset construction}
% \subsection{abstract pattern reasoning task}
% \subsubsection{Human Construction}

\end{document}